\documentclass{article}

\PassOptionsToPackage{numbers,sort&compress}{natbib}
\usepackage[preprint]{neurips_2026}

\usepackage[utf8]{inputenc}
\usepackage[T1]{fontenc}
\usepackage{hyperref}
\hypersetup{hidelinks}
\usepackage{url}
\usepackage{microtype}
\usepackage{amsfonts}

\usepackage{amssymb}
\usepackage{amsmath}
\usepackage{capt-of}
\usepackage{placeins}
\usepackage{booktabs}
\usepackage{graphicx}  
\usepackage{booktabs}  
\usepackage{multirow}
\usepackage{array}
\usepackage{booktabs}
\usepackage{multirow}
\usepackage{graphicx}
\usepackage[table]{xcolor}
\definecolor{blockgray}{gray}{0.93}


\usepackage{pifont}

\begin{document}

\title{An Efficient Training Paradigm for Image Restoration via Improved Fr\'echet Loss}

\author{%
Shanyuan Jie\textsuperscript{1,2,5,\ensuremath{\dagger}} \quad
Wu Shi\textsuperscript{1,\ensuremath{\dagger}} \quad
Xinqi Lin\textsuperscript{3} \quad
Kainan Yan\textsuperscript{1,2} \quad
Zian Wang\textsuperscript{1,2} \\
\bfseries
Yingxue Pang\textsuperscript{5} \quad
Yan Wang\textsuperscript{5} \quad
Shijie Zhao\textsuperscript{5,*} \quad
Chao Dong\textsuperscript{1,4,*} \\
\normalfont
\textsuperscript{1}Shenzhen Institutes of Advanced Technology, CAS \\
\textsuperscript{2}University of Chinese Academy of Sciences \\
\textsuperscript{3}The Chinese University of Hong Kong \\
\textsuperscript{4}Shenzhen University of Advanced Technology \\
\textsuperscript{5}ByteDance Inc.
}

\maketitle

\begingroup
\renewcommand{\thefootnote}{}
\footnotetext{%
\textsuperscript{\ensuremath{\dagger}}Equal contribution. \quad
\textsuperscript{*}Corresponding authors.%
}
\endgroup

\begin{center}
    \includegraphics[
        width=\linewidth,
        keepaspectratio
    ]{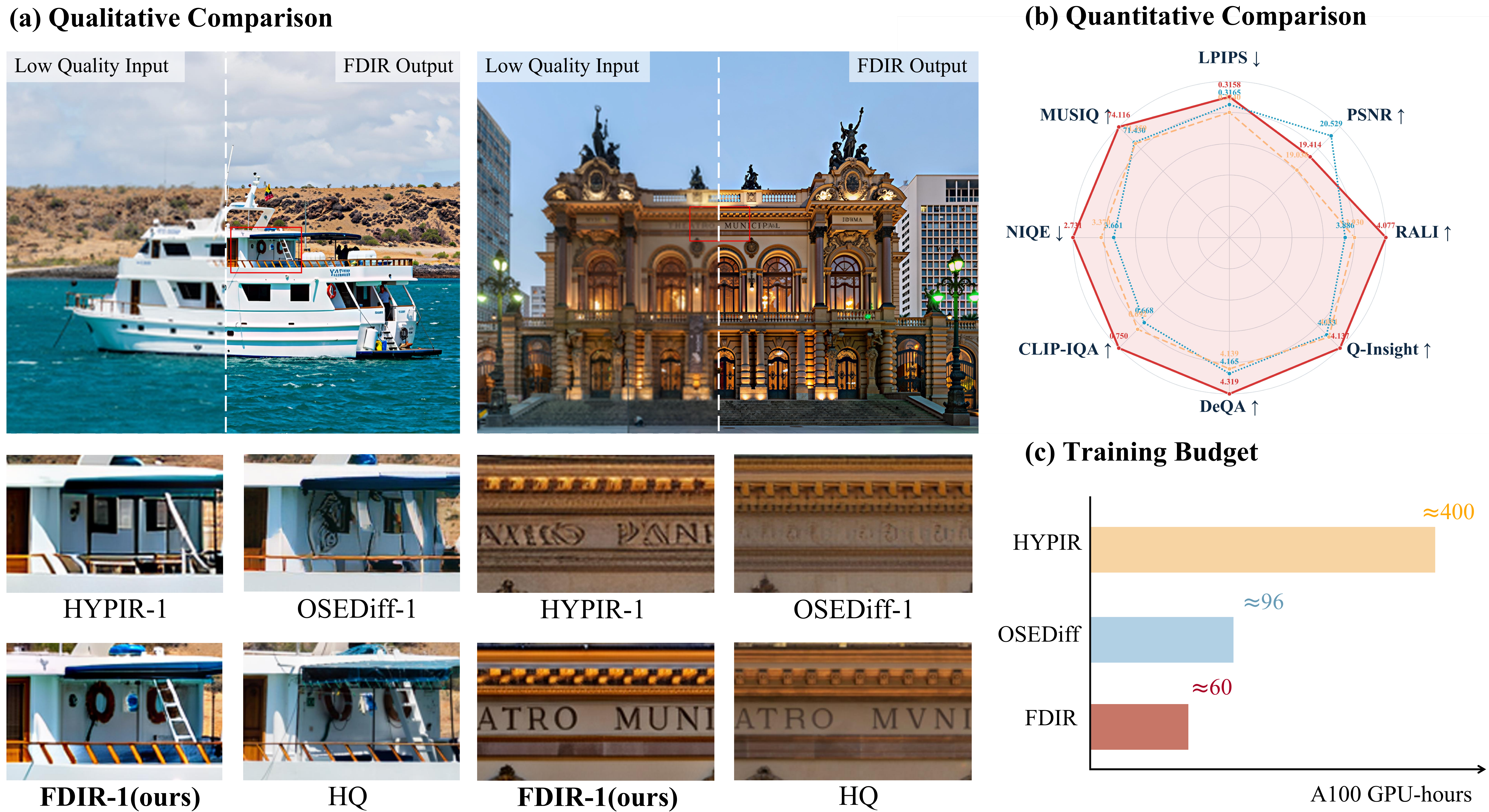}

    \captionof{figure}{
        (a) Qualitative comparisons with representative methods.
        (b) Quantitative comparison on DIV2K-Val using
        full-reference and no-reference metrics.
        (c) Training budgets in A100 GPU-hours.
        OSEDiff uses the officially reported budget,
        while HYPIR is measured under its official schedule.
    }
    \label{fig:teaser}
\end{center}

\begin{abstract}
Diffusion-based image restoration achieves strong perceptual quality but typically relies on iterative sampling, leading to slow inference. Existing single-step methods often use diffusion distillation or adversarial training, complicating optimization. 
To simplify training, we propose FDIR, a framework for training single-pass generative restoration models through regression-style supervised optimization using directly computable objectives.
To realize this paradigm, FDIR employs the Fr\'echet distance
to match the feature distributions of restored and real images
in frozen representation spaces, providing explicit
distribution-level supervision. To better accommodate image restoration, we incorporate
local patch features to capture fine-grained restoration
discrepancies. Moreover, we introduce instance-adaptive
statistics by combining the feature moments of each
SR--HQ pair with synchronized historical statistics,
allowing distribution matching to incorporate
input-specific correspondence while retaining
distribution-level context.
FDIR optimizes only the LoRA parameters of the restoration backbone. Experiments demonstrate strong perceptual restoration quality and efficient training, suggesting a simpler paradigm for generative image restoration.
\end{abstract}









\section{Introduction}

In recent years, diffusion-based image restoration has made remarkable progress by leveraging powerful generative priors, enabling the recovery of highly realistic details under complex degradations~\cite{Diffbir,SUPIR}.
However, the iterative sampling process of diffusion models leads to slow inference, limiting the practical deployment of these methods.

To improve inference efficiency, existing methods employ diffusion distillation~\cite{OSEDiff,TSDSR} or adversarial training~\cite{HYPIR,ODTSR} to perform restoration in a single forward pass. From a distribution-matching perspective, adversarial training and score-based distillation both seek to align generated and target image distributions, but obtain their training signals through different mechanisms. Adversarial methods rely on discriminator-based optimization, whereas score-based distillation methods use diffusion-derived score guidance, often involving additional learned estimators. Although effective, these auxiliary estimation or optimization mechanisms make training more involved.

These limitations motivate us to seek a simpler and more efficient training paradigm for single-pass image restoration. Specifically, we investigate whether generative restoration can be trained through regression-style supervised optimization, in which the restoration model is optimized directly from paired training data using computable objectives, without relying on auxiliary optimization processes or online learned estimators to construct the training signal.

Within this framework, we seek a directly computable objective that also provides distribution-level supervision for generative restoration. Fr\'echet loss (FD-Loss) offers a simple and explicit objective for this purpose by comparing feature statistics of restored and real images in frozen representation spaces~\cite{FD_loss}. However, directly applying conventional FD-Loss to image restoration is non-trivial.


One limitation is that conventional FD-Loss typically performs distribution matching using image-level features. In image restoration, however, the LQ input already strongly constrains image content and spatial layout, while important restoration discrepancies often occur in local textures and high-frequency regions. Such localized errors may be overlooked by global feature aggregation. We therefore extend image-level matching with local FD matching using patch-level features to better capture restoration details.

More fundamentally, conventional FD-Loss performs marginal distribution alignment. As illustrated in Fig.~\ref{fig:method}(b), it matches generated statistics against fixed dataset-level real statistics, without explicitly considering the correspondence between each restored output and its paired HQ target. In image restoration, however, the LQ input specifies the underlying content, making such input-specific correspondence important.

To better incorporate this correspondence, we introduce instance-adaptive statistics into FD matching, as illustrated in Fig.~\ref{fig:method}(c). Specifically, we combine synchronized historical SR/HQ statistics with the feature moments of each current SR output and its paired HQ target. This allows distribution matching to incorporate pair-specific information while retaining stable distribution-level context.

Based on these observations, we present \textbf{FDIR}, a simple and efficient framework for training single-pass generative restoration models through regression-style supervised optimization. FDIR employs FD-based distribution supervision with local feature matching and instance-adaptive statistics, without introducing additional trainable discriminators or diffusion-derived score/trajectory supervision. By optimizing only lightweight restoration adapters, FDIR achieves strong perceptual restoration quality with a lower training budget than representative methods, as illustrated in Fig.~\ref{fig:teaser}.

More broadly, this work represents an initial exploration of training generative restoration models through regression-style supervised optimization. While FD-based supervision provides a simple alternative to existing training mechanisms, its reliance on low-order feature statistics may still leave subtle local texture discrepancies insufficiently constrained. We hope this exploration will encourage further research toward simpler and more effective training strategies, ultimately improving both the efficiency and quality of generative image restoration.








\section{Related Work}

\subsection{Generative Image Restoration}

Image restoration aims to recover high-quality images from low-quality observations affected by unknown and complex degradations~\cite{RealSR,DrealSR}. Early deep learning-based methods primarily focused on reconstruction fidelity, learning deterministic mappings under predefined synthetic degradation settings~\cite{SwinIR}. Although effective under controlled conditions, these methods often struggle with complex real-world degradations and produce over-smoothed details.

To improve perceptual quality, generative image restoration methods introduced adversarial learning to synthesize more realistic textures and fine-grained details~\cite{ESRGAN}. Subsequent studies increasingly focused on real-world restoration, with Real-ESRGAN~\cite{RealESRGAN}
exploring more realistic degradation models to improve
generalization to practical scenarios. Despite their ability to generate visually plausible details, GAN-based approaches may introduce unnatural textures and perceptual artifacts, making it challenging to balance perceptual quality and reconstruction fidelity~\cite{DeSRA,LDL}.

Benefiting from the strong generative priors of diffusion models, diffusion-based restoration methods have demonstrated remarkable capabilities in recovering realistic structures and fine-grained details. StableSR~\cite{StableSR} exploits pretrained text-to-image diffusion priors for real-world super-resolution, while DiffBIR~\cite{Diffbir}, SeeSR~\cite{SeeSR}, and SUPIR~\cite{SUPIR} further explore blind image restoration, semantic guidance, and large-scale generative priors, respectively. AdaptBIR~\cite{AdaptBIR} further adapts latent diffusion priors to improve restoration fidelity. More recently, DiT4SR~\cite{DiT4SR} explores diffusion transformer architectures for real-world image super-resolution. Beyond restoration architectures, UARE~\cite{UARE} investigates unified vision-language-based image quality assessment and restoration, while OARS~\cite{OARS} explores process-aware online alignment for generative real-world image super-resolution. Despite these advances, diffusion- and flow-matching-based restoration methods often rely on iterative sampling, resulting in considerable inference costs.

\subsection{Training One-Step Diffusion Restoration Models}

Existing one-step diffusion restoration methods mainly follow two training paradigms: diffusion distillation~\cite{PiSASR,GenDR,EliminatingVAE} and adversarial optimization~\cite{HYPIR,OMGSR}.

Diffusion-distillation methods transfer pretrained diffusion priors to one-step restoration models through trajectory or score supervision. Trajectory-based methods, such as SinSR~\cite{SinSR}, learn direct one-step mappings from diffusion trajectories or consistency constraints, whereas score distillation methods, including OSEDiff~\cite{OSEDiff}, TSD-SR~\cite{TSDSR}, and PiSA-SR~\cite{PiSASR}, optimize restored outputs using score guidance from pretrained diffusion models. Despite their effectiveness, these methods generally rely on explicitly constructed diffusion trajectories or additional score supervision, increasing training complexity.

Another line of work exploits the generative priors of pretrained diffusion models through adversarial optimization for one-step restoration. HYPIR~\cite{HYPIR} and OMGSR~\cite{OMGSR} fine-tune pretrained diffusion models with perceptual and adversarial supervision. These methods avoid explicit diffusion distillation but require additional discriminators and adversarial optimization, increasing training complexity and computational cost.

Beyond these two paradigms, Representation Fr\'echet Loss~\cite{FD_loss} demonstrates that generative models can be trained by directly matching real and generated feature distributions in frozen representation spaces. Motivated by this formulation, we investigate Fr\'echet-based distribution matching as an alternative training paradigm for one-step diffusion-based image restoration.



\section{Preliminaries}

\paragraph{Fr\'echet Loss}


Fr\'echet Distance (FD)~\cite{FID,FD} measures the
distributional distance between real and generated images
in the feature space of a pretrained encoder $\phi$.

Specifically, let $p_R$ and $p_G$ denote the distributions
of real and generated images, respectively. Their feature
distributions induced by $\phi$ are approximated as
multivariate Gaussians, characterized by their means
and covariances:
\begin{equation}
\begin{aligned}
\boldsymbol{\mu}_{R}
&= \mathbb{E}_{x\sim p_R}[\phi(x)],
&
\boldsymbol{\Sigma}_{R}
&= \operatorname{Cov}_{x\sim p_R}[\phi(x)],\\
\boldsymbol{\mu}_{G}
&= \mathbb{E}_{\hat{x}\sim p_G}[\phi(\hat{x})],
&
\boldsymbol{\Sigma}_{G}
&= \operatorname{Cov}_{\hat{x}\sim p_G}[\phi(\hat{x})].
\end{aligned}
\end{equation}
The resulting Gaussian approximations are denoted by
$\mathcal{R}=(\boldsymbol{\mu}_{R},\boldsymbol{\Sigma}_{R})$
and
$\mathcal{G}=(\boldsymbol{\mu}_{G},\boldsymbol{\Sigma}_{G})$.

The FD between the two Gaussian distributions is defined as
\begin{equation}
\begin{aligned}
\operatorname{FD}(\mathcal R,\mathcal G)
={}&
\left\|
\boldsymbol{\mu}_{R}
-
\boldsymbol{\mu}_{G}
\right\|_2^2
\\
&+
\operatorname{Tr}\left(
\boldsymbol{\Sigma}_{R}
+
\boldsymbol{\Sigma}_{G}
-
2
\left[
\boldsymbol{\Sigma}_{R}
\boldsymbol{\Sigma}_{G}
\right]^{1/2}
\right).
\end{aligned}
\label{eq:frechet_distance}
\end{equation}
Reliable estimation of these statistics requires a sufficiently large number of samples. 
For practical training FD-Loss~\cite{FD_loss} precomputes the real feature statistics $\mathcal R$ over the full real-image dataset. 
The generated feature statistics $\mathcal G$ are estimated online by combining the current generated mini-batch with historical statistics accumulated using exponential moving averages.



To normalize the scale of the FD training objective
across representation spaces,
FD-Loss adopts the following normalized formulation: 
\begin{equation}
\mathcal{L}_{\mathrm{FD}}^{\phi}
(\mathcal R,\mathcal G)
=
\frac{
\operatorname{FD}(\mathcal R,\mathcal G)
}{
\operatorname{sg}\!\left[
\operatorname{FD}(\mathcal R,\mathcal G)
\right]
+\epsilon
},
\label{eq:fd_training_objective}
\end{equation}
where $\operatorname{sg}[\cdot]$ denotes the stop-gradient operation.

\begin{figure*}[t!]
    \centering
    \includegraphics[width=1\linewidth]{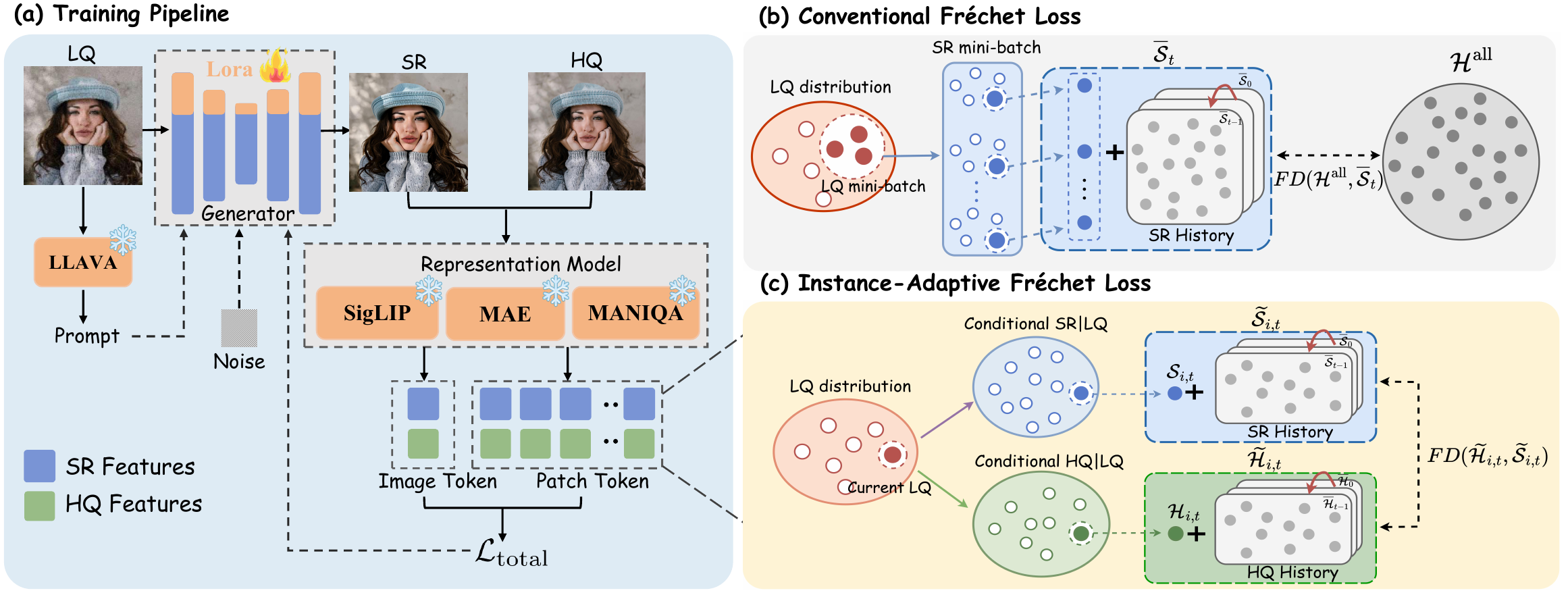}
   \caption{
Overview of FDIR.
(a) Training pipeline of FDIR.
(b) Conventional FD-Loss matches online SR statistics against fixed
dataset-level HQ statistics.
(c) Instance-adaptive FD-Loss is computed using statistics constructed
from the current paired SR--HQ observations and synchronized histories.
}
    \label{fig:method}
\end{figure*}

\section{Method}




We propose \textbf{FDIR}, an efficient framework for direct supervised
training of one-step image restoration models.
As shown in Fig.~\ref{fig:method}(a), given an LQ image and sampled noise,
the restoration model $G_\theta$ produces an SR output in a single forward pass.
The SR output and its paired HQ target are then processed by frozen
representation encoders to compute an image-restoration-oriented FD-Loss.
During training, all encoders remain fixed, and only the LoRA parameters
of $G_\theta$ are optimized.

Specifically, the objective aligns SR and HQ feature distributions at both
image and patch levels to capture global image properties and local restoration details.
It further combines the current SR--HQ pair with synchronized historical
statistics to construct instance-adaptive statistics for FD matching.
The resulting losses across multiple representation spaces are aggregated
to form the overall distribution-matching objective.



\subsection{Locality-Aware Fr\'echet Loss}



Image restoration requires the model to recover a high-fidelity output with realistic details. 
However, FD-Loss relies on global average pooling (in CNNs) or \texttt{[CLS]} tokens (in ViTs), which is largely \textbf{spatially agnostic}. 
In image restoration, LQ inputs usually preserve the global image
content and spatial layout, whereas perceptual
discrepancies are often found around edges, textures,
and high-frequency regions.
Image-level representations can capture holistic properties but overlook such local differences. 
We therefore complement image-level FD matching with
patch-level FD matching to jointly capture global
structure and local restoration details.

Specifically, given a frozen encoder $\phi$, we extract
an image-level feature
$\phi^{\mathrm{img}}(x)\in\mathbb{R}^{D_\phi}$
and patch-level features
$\phi^{\mathrm{pat}}(x)\in
\mathbb{R}^{N_\phi\times D_\phi}$
from an image $x$. The former is obtained from the
\texttt{[CLS]} token or global pooling, while the latter
retains spatial tokens as individual samples for
distribution estimation.

Let $\mathcal{S}$ and $\mathcal{H}$ denote the Gaussian
feature statistics of restored images (SR) and their
paired high-quality targets (HQ), respectively. Their
image-level and patch-level statistics are given by
$\mathcal{S}^{\mathrm{img/pat}}=
(\boldsymbol{\mu}_{\mathrm{SR}}^{\mathrm{img/pat}},
\boldsymbol{\Sigma}_{\mathrm{SR}}^{\mathrm{img/pat}})$
and
$\mathcal{H}^{\mathrm{img/pat}}=
(\boldsymbol{\mu}_{\mathrm{HQ}}^{\mathrm{img/pat}},
\boldsymbol{\Sigma}_{\mathrm{HQ}}^{\mathrm{img/pat}})$.

The corresponding image-level FD and patch-level FD
(\textbf{PatchFD}) objectives are defined as follows:
\begin{equation}
\begin{aligned}
\mathcal{L}_{\mathrm{FD}}^{\phi,\mathrm{img}}
&=
\mathcal{L}_{\mathrm{FD}}^{\phi}
\left(
\mathcal{H}^{\mathrm{img}},\mathcal{S}^{\mathrm{img}}
\right),\\
\mathcal{L}_{\mathrm{FD}}^{\phi,\mathrm{pat}}
&=
\mathcal{L}_{\mathrm{FD}}^{\phi}
\left(
\mathcal{H}^{\mathrm{pat}},\mathcal{S}^{\mathrm{pat}}
\right).
\end{aligned}
\label{eq:patch_fd}
\end{equation}

\begin{figure}[h!]
    \centering
    \includegraphics[width=0.75\linewidth]{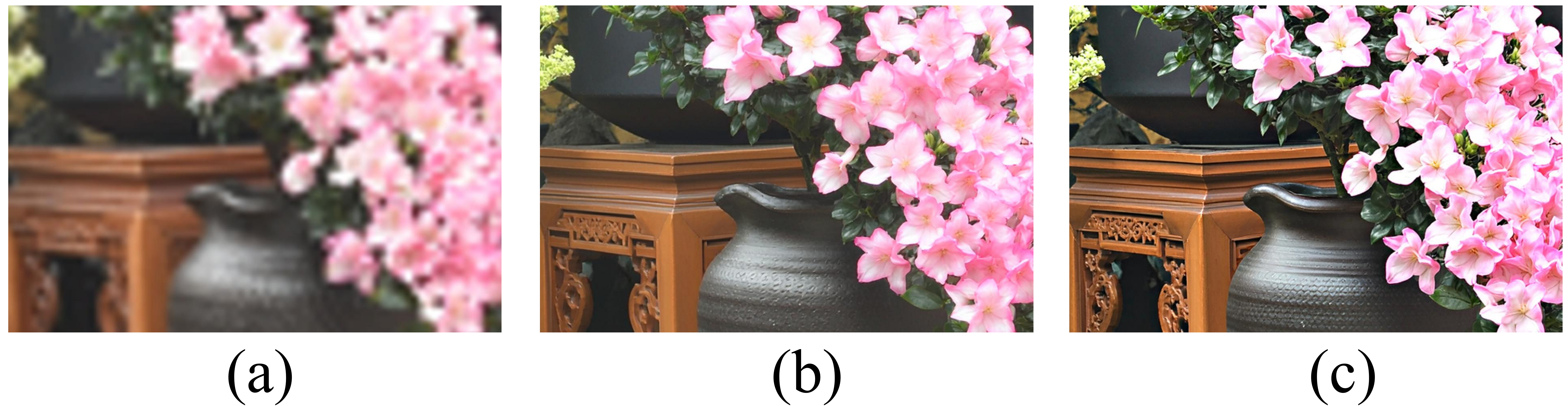}
    \caption{
(a) LQ input. (b) Restoration using image-level FD only.
(c) Restoration using both image-level and patch-level FD.
Patch-level FD improves local texture.}
    \label{fig:LAF_method}
\end{figure}

%
As illustrated in Fig.~\ref{fig:LAF_method}, patch-level
matching provides additional supervision for localized
restoration details. 


\subsection{Instance-Adaptive Statistics}

\paragraph{Limitations of Conventional FD-Loss}


Image restoration is fundamentally a conditional problem. 
As shown in Fig.~\ref{fig:method} (b), the conventional FD-Loss combines the feature statistics of the current SR mini-batch with historical SR statistics to obtain the online SR statistics $\overline{\mathcal S}_{t}$, and matches them against the fixed full-dataset HQ statistics $\mathcal H^{\mathrm{all}}$ through $\mathcal L_{\mathrm{FD}}^{\phi} (\mathcal H^{\mathrm{all}},\overline{\mathcal S}_{t})$.
Ideally, the model should match the statistics between the SR output and the HQ target given the LQ observation. 
The fixed global HQ reference does not explicitly
encode the correspondence between each LQ input
and its paired HQ target.
Consequently, the conventional FD-Loss only matches the SR results with the statistics of marginal distributions of HQ images.

\paragraph{LQ-Conditioned Restoration Target}
As shown in Fig.~\ref{fig:method} (c), given an LQ input $x_{\mathrm{LQ},i}$, image restoration ideally requires the distribution of restored outputs to match the HQ distribution conditioned on the same observation. 
This objective can be expressed as
\begin{equation}
\operatorname{FD}
\left(
\mathcal{H}(x_{\mathrm{HQ}} | x_{\mathrm{LQ},i}),
\mathcal{S}(x_{\mathrm{SR}} | x_{\mathrm{LQ},i})
\right),
\label{eq:lq_conditioned_fd}
\end{equation}
where $\mathcal{H}(x_{\mathrm{HQ}} | x_{\mathrm{LQ},i})$ and $\mathcal{S}(x_{\mathrm{SR}} | x_{\mathrm{LQ},i})$ denote the HQ and SR feature statistics conditioned on the same LQ.

Directly estimating such LQ-conditioned statistics is impractical, since each LQ condition typically provides only a single paired observation, which is insufficient for reliably estimating the feature statistics required by the FD-Loss. 
We therefore combine the current SR--HQ pair with synchronized historical statistics to construct practical instance-adaptive statistics for FD-Loss that incorporates pair-specific information while retaining stable distribution-level context.

\paragraph{Instance-Adaptive Fr\'echet Loss}




During training, we keep track of the statistics of both HQ and SR features using EMA-based estimation. 
Specifically, we maintain exponential moving average of the first and second moments with coefficient $\beta \in (0, 1)$. 
At the end of training iteration \(t\), the EMA computes
\begin{equation}
\begin{aligned}
    \bar{\mu}_{\mathrm{HQ/SR}}^{(t)}
    &= \beta \bar{\mu}_{\mathrm{HQ/SR}}^{(t-1)}
    + (1-\beta)\mu_{\mathrm{HQ/SR}}^{\mathrm{batch}(t)}, \\
    \bar{M}_{\mathrm{HQ/SR}}^{(t)}
    &= \beta \bar{M}_{\mathrm{HQ/SR}}^{(t-1)}
    + (1-\beta)M_{\mathrm{HQ/SR}}^{\mathrm{batch}(t)},
\end{aligned}
\label{eq:ema_moments}
\end{equation}
where $\mu_{\mathrm{HQ/SR}}^{\mathrm{batch}(t)}$ and $M_{\mathrm{HQ/SR}}^{\mathrm{batch}(t)}$ are the first and second moments of the current batch of HQ and SR features $\{\phi(x_{\mathrm{HQ}, i}^{(t)})\}$ and $\{\phi(x_{\mathrm{SR}, i}^{(t)})\}$.
The online EMA estimators provide synchronized historical HQ and SR
statistics, denoted by $\overline{\mathcal H}_{t}$ and
$\overline{\mathcal S}_{t}$, respectively. These statistics are
represented by the corresponding EMA first and second raw moments in
Eq.~\eqref{eq:ema_moments} and are synchronized because they are
updated from the same sequence of paired training batches.

To construct the instance-adaptive FD-Loss, for the $i$-th paired
sample, we linearly interpolate its current feature moments with the
corresponding pre-update historical moments:
\begin{equation}
\begin{aligned}
    \widetilde{\mu}_{\mathrm{HQ/SR}}^{(i,t)}
    &= \alpha \bar{\mu}_{\mathrm{HQ/SR}}^{(t-1)}
    + (1-\alpha)\mu_{\mathrm{HQ/SR}}^{(i,t)}, \\
    \widetilde{M}_{\mathrm{HQ/SR}}^{(i,t)}
    &= \alpha \bar{M}_{\mathrm{HQ/SR}}^{(t-1)}
    + (1-\alpha)M_{\mathrm{HQ/SR}}^{(i,t)},
\end{aligned}
\label{eq:ada_ema_moments}
\end{equation}
Here, $\alpha$ controls the contribution of the historical statistics,
while $1-\alpha$ controls that of the current pair.
$\boldsymbol{\mu}_{\mathrm{HQ/SR}}^{(i,t)}$ and
$\mathbf M_{\mathrm{HQ/SR}}^{(i,t)}$ denote the first and second raw
moments of the current HQ/SR instance, respectively.

For image-level representations, each instance provides a single feature
vector $\mathbf f$, such that
$\boldsymbol{\mu}^{(i,t)}=\mathbf f$ and
$\mathbf M^{(i,t)}=\mathbf f\mathbf f^{\top}$; for patch-level
representations, these moments are computed over the spatial tokens.
In both cases, the covariance is recovered only after the current
moments are interpolated with the historical EMA moments, rather than
being estimated from the current instance alone. 
The corresponding covariance is then given by
\begin{equation}
\widetilde{\boldsymbol{\Sigma}}_{\mathrm{HQ/SR}}^{(i,t)}
=
\widetilde{\mathbf M}_{\mathrm{HQ/SR}}^{(i,t)}
-
\widetilde{\boldsymbol{\mu}}_{\mathrm{HQ/SR}}^{(i,t)}
\widetilde{\boldsymbol{\mu}}_{\mathrm{HQ/SR}}^{(i,t)\top}.
\label{eq:adafd_covariance}
\end{equation}

Accordingly, the 
instance-adaptive HQ and SR statistics are
defined as
\begin{equation}
\begin{aligned}
\widetilde{\mathcal H}_{i,t}
&=
\left(
\widetilde{\boldsymbol{\mu}}_{\mathrm{HQ}}^{(i,t)},
\widetilde{\boldsymbol{\Sigma}}_{\mathrm{HQ}}^{(i,t)}
\right),\\
\widetilde{\mathcal S}_{i,t}
&=
\left(
\widetilde{\boldsymbol{\mu}}_{\mathrm{SR}}^{(i,t)},
\widetilde{\boldsymbol{\Sigma}}_{\mathrm{SR}}^{(i,t)}
\right).
\end{aligned}
\label{eq:adafd_statistics}
\end{equation}
As illustrated in Fig.~\ref{fig:method}(c), these statistics combine
pair-specific information with synchronized historical context.
Using these statistics at each feature level, we compute
the instance-adaptive Fr\'echet loss (\textbf{AdaFD}) as
%
\begin{equation}
\mathcal L_{\mathrm{AdaFD}}^{\phi,\ell}
=
\frac{1}{B}
\sum_{i=1}^{B}
\mathcal L_{\mathrm{FD}}^{\phi}
\left(
\widetilde{\mathcal H}_{i,t},
\widetilde{\mathcal S}_{i,t}
\right),
\quad \ell\in\{\mathrm{img},\mathrm{pat}\},
\label{eq:adafd_objective}
\end{equation}

where $B$ denotes the mini-batch size. All samples in the current
mini-batch use the same pre-update historical statistics. Gradients are
propagated only through the current SR features, while the paired HQ
features and historical statistics are treated as fixed.

\subsection{Overall Training Objective}
\label{sec:overall_training_objective}

To provide distribution-level supervision across multiple feature
spaces, we compute the instance-adaptive FD-Loss using a set of frozen
representation encoders $\Phi$. For each encoder $\phi\in\Phi$ and
feature granularity $\ell\in\{\mathrm{img},\mathrm{pat}\}$, the
corresponding loss term is denoted by
$\mathcal{L}_{\mathrm{AdaFD}}^{\phi,\ell}$.
We aggregate the image-level and patch-level terms across all
representation spaces and denote the resulting complete FD objective by
$\mathcal{L}_{\mathrm{FD}}^{\mathrm{full}}$:

\begin{equation}
\mathcal{L}_{\mathrm{FD}}^{\mathrm{full}}
=
\frac{1}{|\Phi|}
\sum_{\phi\in\Phi}
\left(
\mathcal{L}_{\mathrm{AdaFD}}^{\phi,\mathrm{img}}
+
\mathcal{L}_{\mathrm{AdaFD}}^{\phi,\mathrm{pat}}
\right).
\label{eq:full_fd_objective}
\end{equation}

The complete training objective of FDIR combines reconstruction,
perceptual, and distribution-level supervision:

\begin{equation}
\mathcal{L}_{\mathrm{total}}
=
\lambda_{\mathrm{MSE}}\mathcal{L}_{\mathrm{MSE}}
+
\lambda_{\mathrm{LPIPS}}\mathcal{L}_{\mathrm{LPIPS}}
+
\lambda_{\mathrm{FD}}
\mathcal{L}_{\mathrm{FD}}^{\mathrm{full}}.
\label{eq:overall_training_objective}
\end{equation}



\section{Experiments}

\subsection{Experimental Setup}
\label{sec:experimental_setup}

\begin{figure}[t]
    \centering
    \includegraphics[width=\linewidth]{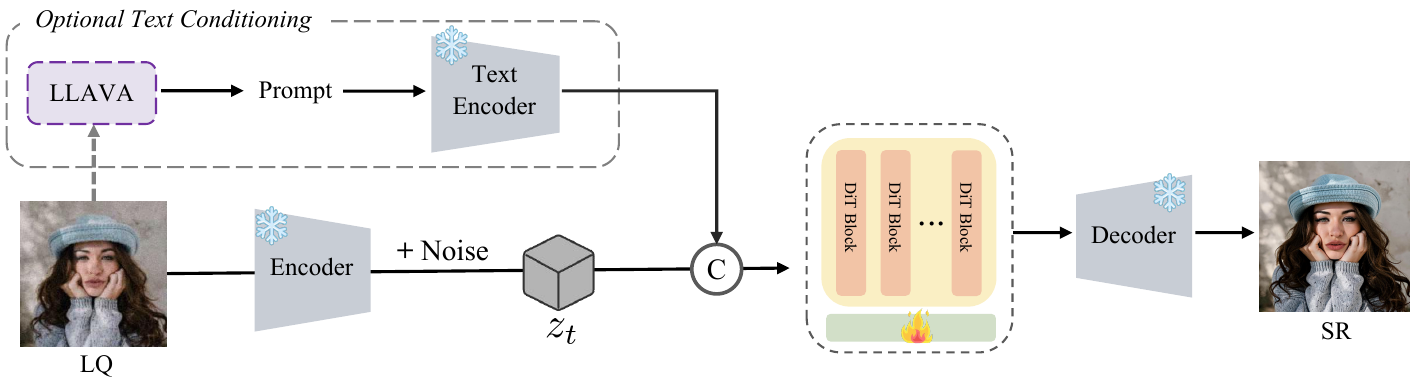}
    \caption{
    Overview of the single-step restoration generator. The
    noise-augmented LQ latent and textual condition are processed by
    the LoRA-adapted FLUX.2 backbone for single-step restoration.}
    \label{fig:restoration_generator}
\end{figure}

\paragraph{Model Configuration}
We adopt FLUX.2-klein-base-4B as the pretrained generative backbone
and optimize only its LoRA adapters for single-step image restoration.
FDIR employs SigLIP, MAE, and MANIQA as the semantic,
reconstruction-oriented, and quality-aware representation encoders,
respectively. We set the LoRA rank to 64 and optimize the adapters
with Adam using a learning rate of $1\times10^{-5}$ and a global
batch size of 64. As illustrated in Fig.~\ref{fig:restoration_generator},
the restoration generator supports optional textual conditioning and
controllable latent noise augmentation. Detailed settings are provided in
Appendix~\ref{app:implementation}.

\paragraph{Training and Data}
We train FDIR for 5K iterations on $8\times$ NVIDIA A100 GPUs.
The training set contains 200K high-quality
images, and paired LQ inputs are synthesized using the two-stage
Real-ESRGAN degradation pipeline~\cite{RealESRGAN}.

\paragraph{Evaluation Datasets and Metrics}
We evaluate FDIR on the synthetic paired DIV2K-Val benchmark~\cite{Div2k}
and the real-world unpaired RealPhoto60~\cite{SUPIR} and
RealLQ250~\cite{RealLQ250} benchmarks. For DIV2K-Val, we report PSNR,
SSIM, LPIPS, and DISTS. For the real-world benchmarks, we report MUSIQ,
NIQE, CLIP-IQA, DeQA, Q-Insight, and RALI.

\subsection{Comparison with Other Methods}

\paragraph{Comparison Methods}
We compare our method with representative and state-of-the-art
image restoration methods, including the GAN-based method
Real-ESRGAN~\cite{RealESRGAN}; multi-step diffusion-based methods,
including DiffBIR~\cite{Diffbir}, SUPIR~\cite{SUPIR}, and
DiT4SR~\cite{DiT4SR}; and one-step diffusion-based methods,
including OSEDiff~\cite{OSEDiff}, PiSA-SR~\cite{PiSASR},
TSD-SR~\cite{TSDSR}, FiDeSR~\cite{FiDeSR},
ODTSR~\cite{ODTSR}, HYPIR~\cite{HYPIR}, and
OMGSR~\cite{OMGSR} (FLUX-based).

\begin{figure*}[t]
    \centering
    \includegraphics[width=\linewidth]{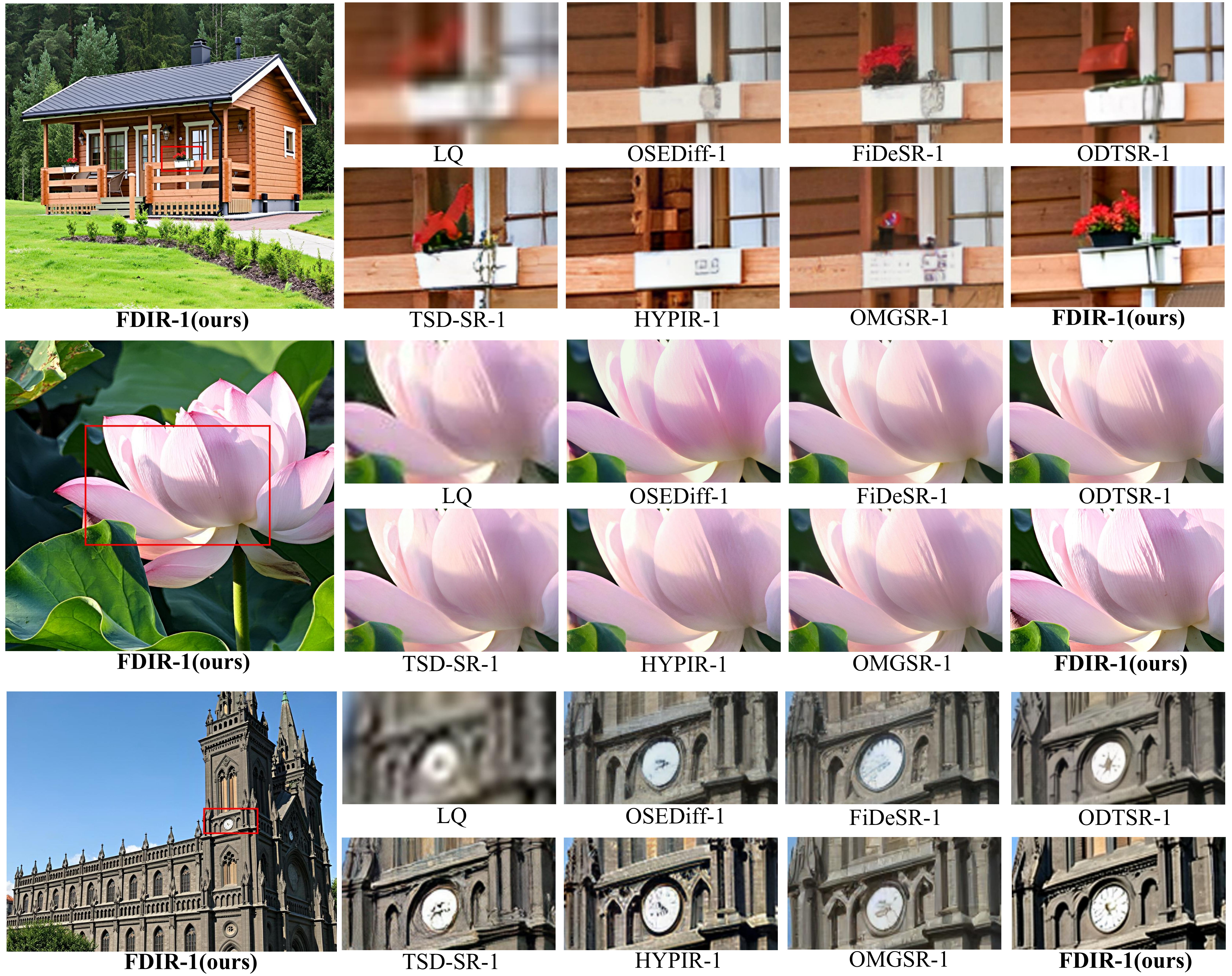}
    \caption{Qualitative comparisons with existing restoration methods on
    the real-world benchmarks RealPhoto60 and RealLQ250.}
    \label{fig:realworld_visual}
\end{figure*}

\begin{figure*}[t]
    \centering
    \includegraphics[width=1\linewidth]{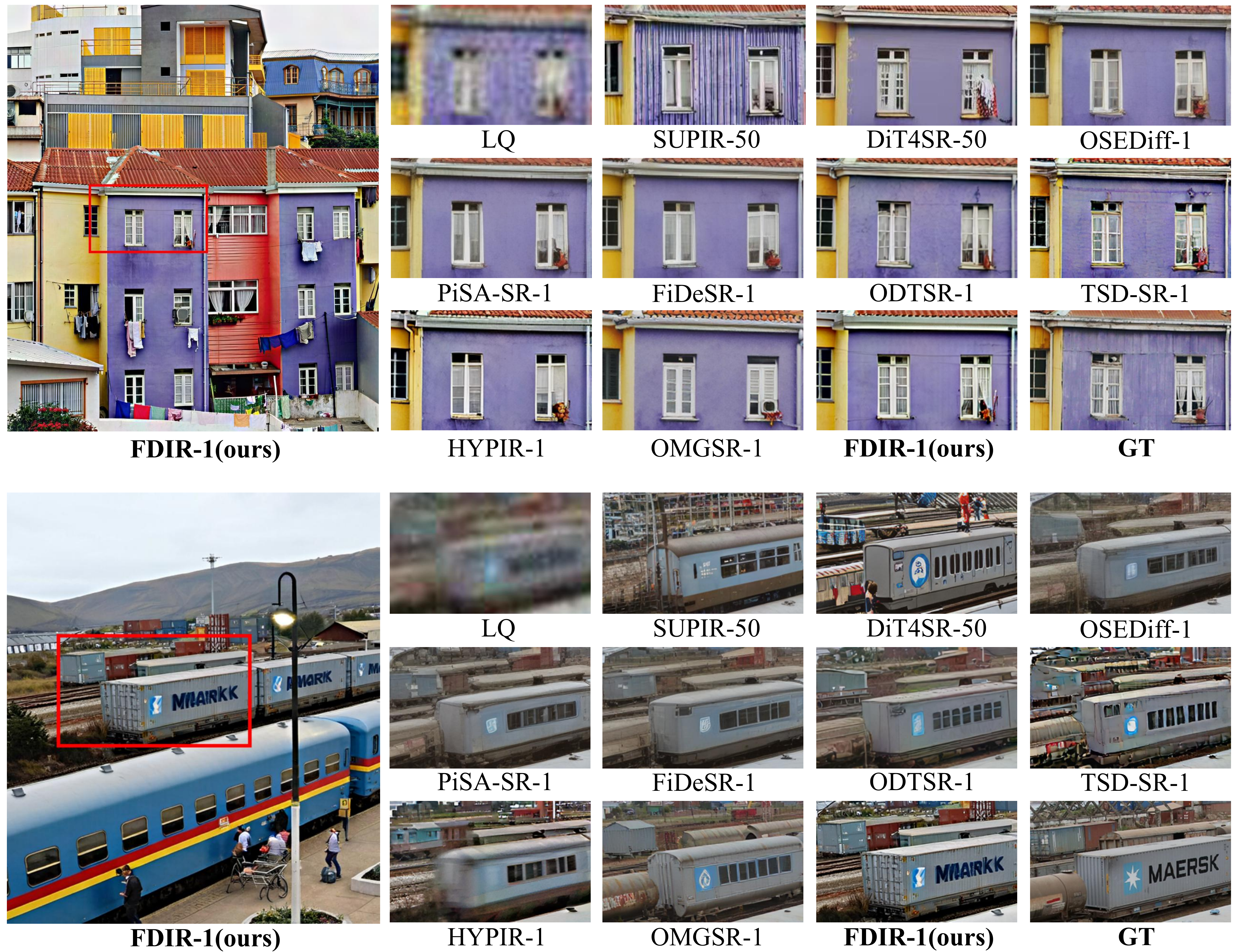}
    \caption{
    Qualitative comparisons with existing
    restoration methods on DIV2K-Val.
}
    \label{fig:div2k_visual}
\end{figure*}

\paragraph{Qualitative Results}
Figure~\ref{fig:realworld_visual} presents qualitative comparisons
on the real-world benchmarks RealPhoto60 and RealLQ250.
Compared with existing methods, FDIR restores clearer
textures and more coherent structures while avoiding
excessive smoothing and conspicuous artifacts.
Figure~\ref{fig:div2k_visual} further presents comparisons
on the synthetic paired DIV2K-Val benchmark, where
FDIR generates realistic details while preserving
the underlying image content.
Overall, these comparisons demonstrate the strong
visual restoration capability of FDIR under different
degradation conditions.

\paragraph{Quantitative Results}
Table~\ref{tab:realphoto_reallq_quantitative} reports quantitative
comparisons on the same two real-world benchmarks, RealPhoto60 and
RealLQ250, while Table~\ref{tab:div2k_quantitative} presents results
on the synthetic paired DIV2K-Val benchmark.
FDIR achieves strong perceptual performance across all three
benchmarks, outperforming competing methods on most
no-reference IQA metrics and recent learned perceptual
evaluators. These results demonstrate the effectiveness
of FDIR in producing high-quality restorations across
different evaluation settings.

\begin{table}[t]
\centering
\caption{
Quantitative comparison on RealPhoto60 and RealLQ250.
C-IQA and Q-Ins. denote CLIP-IQA and Q-Insight, respectively.
Best and second-best results are highlighted in bold and underlined,
respectively.}
\label{tab:realphoto_reallq_quantitative}
\small
\setlength{\tabcolsep}{2pt}
\renewcommand{\arraystretch}{1.12}
\resizebox{\linewidth}{!}{%
\begin{tabular}{@{}l*{12}{c}@{}}
\toprule
\multirow{2}{*}{\textbf{Method}}
& \multicolumn{6}{c}{\textbf{RealPhoto60}}
& \multicolumn{6}{c}{\textbf{RealLQ250}} \\
\cmidrule(lr){2-7}
\cmidrule(lr){8-13}
& MUSIQ$\uparrow$ & NIQE$\downarrow$ & C-IQA$\uparrow$
& DeQA$\uparrow$ & Q-Ins.$\uparrow$ & RALI$\uparrow$
& MUSIQ$\uparrow$ & NIQE$\downarrow$ & C-IQA$\uparrow$
& DeQA$\uparrow$ & Q-Ins.$\uparrow$ & RALI$\uparrow$ \\
\midrule
DiffBIR-50
& 63.536 & 4.697 & 0.716 & 3.784 & 3.773 & 3.684
& 67.528 & 5.119 & 0.714 & 3.876 & 3.829 & 3.750 \\
SUPIR-50
& 71.269 & \underline{3.034} & 0.643 & 4.313 & 4.079 & 3.990
& 67.640 & \underline{3.453} & 0.586 & 4.047 & 3.930 & 3.819 \\
DiT4SR-50
& \underline{74.007} & 3.476 & \underline{0.777}
& 4.281 & \underline{4.117} & \underline{4.125}
& \underline{72.418} & 3.611 & \underline{0.725}
& 4.079 & \underline{4.049} & \underline{4.005} \\
\midrule
Real-ESRGAN-1
& 63.229 & 5.831 & 0.738 & 3.618 & 3.569 & 3.564
& 63.107 & 4.129 & 0.548 & 3.670 & 3.682 & 3.700 \\
OSEDiff-1
& 70.540 & 3.717 & 0.709 & 4.302 & 4.068 & 3.969
& 69.582 & 3.970 & 0.673 & 4.095 & 3.983 & 3.860 \\
PiSA-SR-1
& 72.131 & 3.550 & 0.730 & 4.235 & 4.064 & 3.962
& 71.246 & 3.916 & 0.705 & 4.050 & 3.990 & 3.866 \\
TSD-SR-1
& 73.882 & 3.193 & 0.759 & 4.261 & 4.075 & 4.051
& 72.100 & 3.489 & 0.722 & 3.985 & 3.949 & 3.860 \\
FiDeSR-1
& 72.901 & 3.616 & 0.742 & \underline{4.345}
& \underline{4.117} & 3.999
& 71.653 & 3.898 & 0.707 & \underline{4.163} & 4.023 & 3.908 \\
ODTSR-1
& 70.307 & 3.913 & 0.670 & 4.158 & 4.001 & 3.904
& 69.654 & 4.222 & 0.642 & 3.945 & 3.914 & 3.797 \\
HYPIR-1
& 71.379 & 3.365 & 0.741 & 4.254 & 4.081 & 4.026
& 68.849 & 3.680 & 0.688 & 4.036 & 3.976 & 3.893 \\
OMGSR-1
& 69.242 & 3.623 & 0.677 & 4.306 & 4.074 & 4.021
& 68.732 & 3.985 & 0.644 & 4.154 & 3.976 & 3.923 \\
\midrule
\textbf{FDIR-1 (Ours)}
& \textbf{74.832} & \textbf{2.820} & \textbf{0.788}
& \textbf{4.371} & \textbf{4.195} & \textbf{4.165}
& \textbf{73.601} & \textbf{3.105} & \textbf{0.756}
& \textbf{4.169} & \textbf{4.087} & \textbf{4.046} \\
\bottomrule
\end{tabular}%
}
\end{table}

While FDIR does not achieve the best scores on
full-reference metrics such as PSNR and SSIM on DIV2K-Val, generative restoration models may produce
realistic details that differ from the ground-truth images,
resulting in less favorable reference-based scores despite
improved perceptual quality. As discussed in previous
studies~\cite{TSDSR,SUPIR,HYPIR}, these metrics may not
fully reflect human perceptual preferences. 
Therefore,
we further conduct a user study to assess perceptual
restoration quality and content fidelity beyond
metric-based evaluations.

\begin{table}[t]
\centering
\caption{Quantitative comparison on DIV2K-Val.
Best and second-best results are highlighted in
bold and underlined, respectively.
C-IQA denotes CLIP-IQA, Q-Ins. denotes Q-Insight.}
\small
\setlength{\tabcolsep}{1pt}
\renewcommand{\arraystretch}{1.12}

\begin{tabular*}{\textwidth}{@{\extracolsep{\fill}}l*{10}{c}@{}}
\toprule
\textbf{Method}
& PSNR$\uparrow$
& SSIM$\uparrow$
& LPIPS$\downarrow$
& DISTS$\downarrow$
& MUSIQ$\uparrow$
& NIQE$\downarrow$
& C-IQA$\uparrow$
& DeQA$\uparrow$
& Q-Ins.$\uparrow$
& RALI$\uparrow$ \\
\midrule

DiffBIR-50
& \underline{20.844} & 0.513 & 0.376 & 0.213
& 69.966 & 4.409 & \underline{0.730}
& 3.944 & 3.886 & 3.773 \\

SUPIR-50
& 19.310 & 0.480 & 0.360 & 0.158
& 72.199 & 3.081 & 0.624
& 4.255 & 4.080 & 3.955 \\

DiT4SR-50
& 18.225 & 0.468 & 0.382 & 0.170
& 73.706 & 3.266 & 0.723
& \underline{4.281} & \underline{4.133}
& \textbf{4.081} \\
\midrule

RealESR-1
& 20.781 & \textbf{0.557} & 0.382 & 0.221
& 65.763 & 3.738 & 0.580
& 3.803 & 3.691 & 3.723 \\

OSEDiff-1
& 20.529 & 0.537 & 0.316 & 0.172
& 71.430 & 3.661 & 0.668
& 4.165 & 4.033 & 3.886 \\

PiSA-SR-1
& 20.638 & 0.532 & 0.318 & 0.174
& 72.367 & 3.692 & 0.687
& 4.066 & 4.001 & 3.876 \\

TSD-SR-1
& 18.953 & 0.490 & 0.313 & 0.154
& \underline{73.853} & \underline{3.015} & 0.702
& 4.052 & 3.976 & 3.889 \\

FiDeSR-1
& \textbf{20.988} & \underline{0.553}
& \underline{0.301} & 0.163
& 72.877 & 3.586 & 0.696
& 4.242 & 4.078 & 3.948 \\

ODTSR-1
& 20.408 & 0.527 & 0.335
& \underline{0.146}
& 70.881 & 3.757 & 0.608
& 3.968 & 3.919 & 3.745 \\

HYPIR-1
& 19.038 & 0.507 & 0.324
& \textbf{0.145}
& 71.359 & 3.370 & 0.692
& 4.139 & 4.051 & 3.930 \\

OMGSR-1
& 20.082 & 0.527 & \textbf{0.298}
& \underline{0.146}
& 71.433 & 3.496 & 0.623
& 4.249 & 4.038 & 3.962 \\
\midrule

\textbf{FDIR-1}
& 19.414 & 0.506 & 0.316 & 0.149
& \textbf{74.116} & \textbf{2.721}
& \textbf{0.750} & \textbf{4.319}
& \textbf{4.137} & \underline{4.077} \\
\bottomrule
\end{tabular*}

\label{tab:div2k_quantitative}
\end{table}


\paragraph{User Study}


We conducted a user study to further evaluate the perceptual quality
and content fidelity of different Real-ISR methods. We randomly
sampled 20 real-world LQ images from RealLQ250 and generated their
restored results using FDIR, HYPIR, OMGSR, and ODTSR. We recruited
24 participants, each of whom evaluated all 20 comparison sets.
For each LQ input, the participants selected the best restored result
by jointly considering perceptual quality and consistency with the
input content. The method identities were concealed, and the presentation order was
randomized for each comparison.

\begin{figure}
    \centering
    \includegraphics[width=0.6\linewidth]{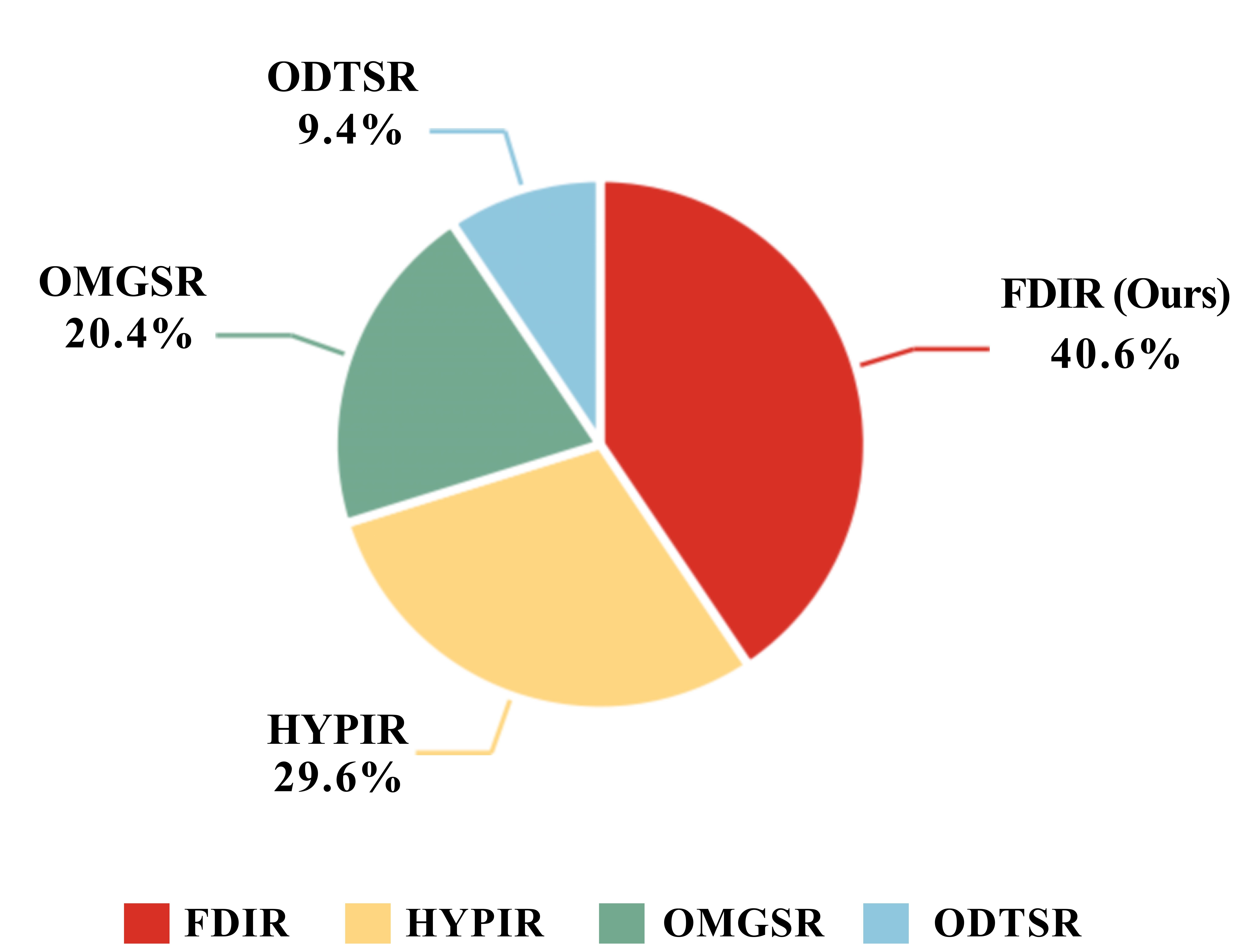}
    \caption{ User-study preference rates for different
   image restoration methods.}
    \label{fig:user_study}
\end{figure}

As shown in Fig.~\ref{fig:user_study}, FDIR achieves the highest
preference rate of 40.6\%, compared with 29.6\%, 20.4\%, and 9.4\%
for HYPIR, OMGSR, and ODTSR, respectively. These results further
demonstrate that FDIR achieves a favorable balance between perceptual
quality and content fidelity.

\begin{figure}[t]
    \centering
    \includegraphics[width=1.0\linewidth]{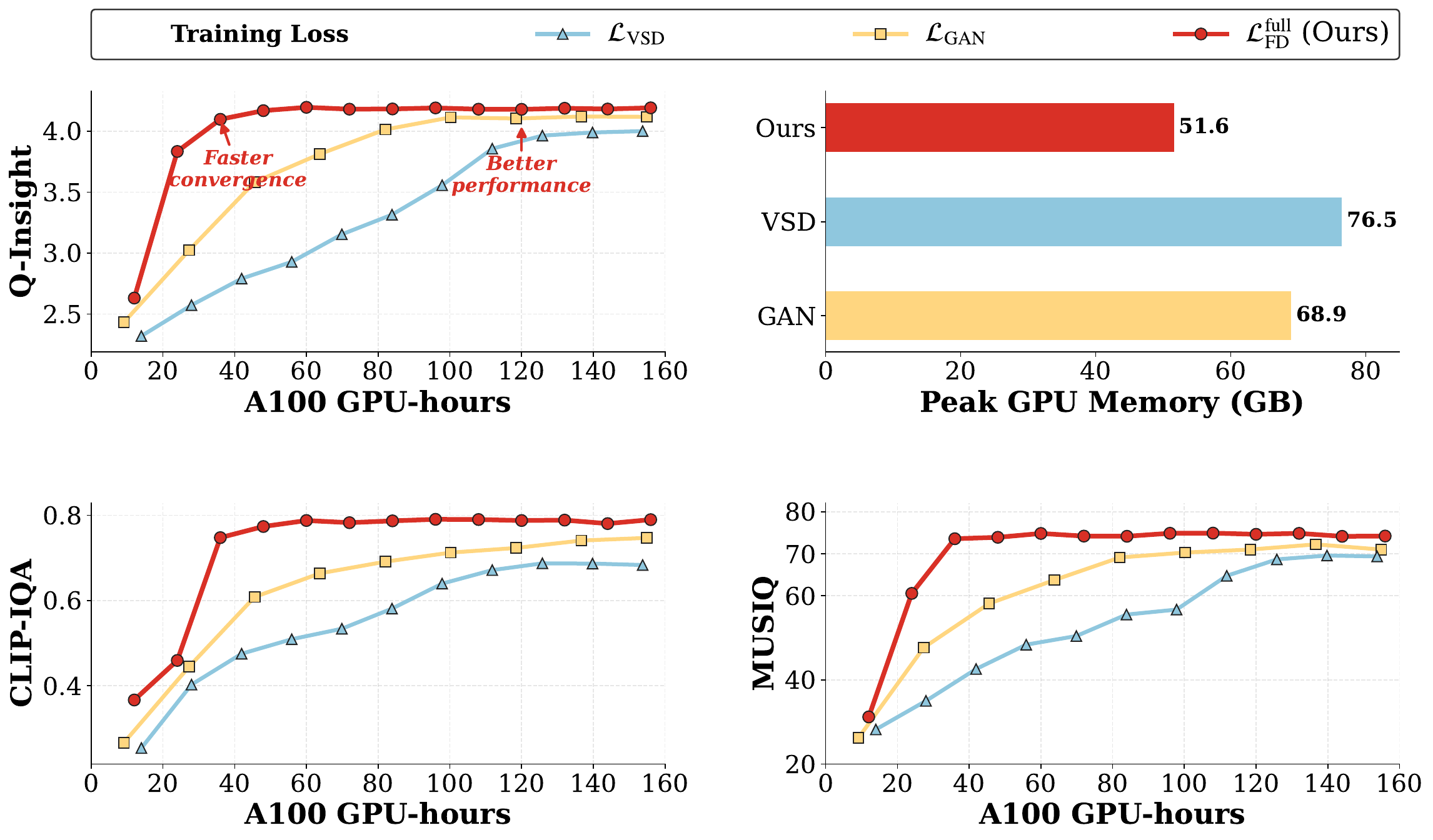}
    \caption{Training efficiency comparison of the full FD,
VSD, and GAN objectives under a shared
restoration framework.}
    \label{fig:loss_efficiency}
\end{figure}

\begin{table}[t]
\centering
\caption{
Restoration quality comparison on RealPhoto60
under approximately matched A100 GPU-hour budgets.
The best results within each budget group are
highlighted in bold.
}
\label{tab:matched_budget}

\setlength{\tabcolsep}{4pt}
\resizebox{\linewidth}{!}{%
\begin{tabular}{cccccccc}
\toprule
GPU-h & Additional Objective & MUSIQ$\uparrow$ & NIQE$\downarrow$
& CLIP-IQA$\uparrow$ & DeQA$\uparrow$
& Q-Insight$\uparrow$ & RALI$\uparrow$ \\
\midrule
\multirow{3}{*}{$\approx 40$}
& $\mathcal{L}_{\mathrm{FD}}^{\mathrm{full}}$
 & \textbf{73.579} & \textbf{3.098}
& \textbf{0.748} & \textbf{4.219}
& \textbf{4.097} & \textbf{4.152} \\
& $\mathcal{L}_{\mathrm{VSD}}$
 & 42.595 & 4.731 & 0.475 & 2.162 & 2.791 & 2.577 \\
& $\mathcal{L}_{\mathrm{GAN}}$
 & 52.089 & 4.595 & 0.569 & 3.448 & 3.252 & 3.672 \\
\midrule
\multirow{3}{*}{$\approx 60$}
& $\mathcal{L}_{\mathrm{FD}}^{\mathrm{full}}$
 & \textbf{74.832} & \textbf{2.820}
& \textbf{0.788} & \textbf{4.371}
& \textbf{4.195} & \textbf{4.165} \\
& $\mathcal{L}_{\mathrm{VSD}}$
 & 48.386 & 4.505 & 0.509 & 2.394 & 2.928 & 2.741 \\
& $\mathcal{L}_{\mathrm{GAN}}$
 & 63.758 & 3.627 & 0.664 & 3.910 & 3.812 & 3.859 \\
\midrule
\multirow{3}{*}{$\approx 150$}
& $\mathcal{L}_{\mathrm{FD}}^{\mathrm{full}}$
 & \textbf{74.214} & \textbf{2.831}
& \textbf{0.790} & \textbf{4.341}
& \textbf{4.191} & \textbf{4.124} \\
& $\mathcal{L}_{\mathrm{VSD}}$
 & 69.370 & 3.912 & 0.683 & 4.130 & 4.000 & 3.949 \\
& $\mathcal{L}_{\mathrm{GAN}}$
 & 72.002 & 3.063
 & 0.767 & 4.284 & 4.117 & 4.044 \\
\bottomrule
\end{tabular}%
}
\end{table}



\subsection{Training Effectiveness and Efficiency}
\label{sec:training_efficiency}

In this section, we evaluate the effectiveness and
training efficiency of FDIR against two widely adopted
training mechanisms for one-step diffusion restoration:
score distillation and adversarial training.
We select OSEDiff~\cite{OSEDiff} and
HYPIR~\cite{HYPIR} as representative approaches
for variational score distillation (VSD) and
adversarial training (GAN), respectively.

For a controlled comparison, we adapt the respective
training objectives to the same FLUX.2-based
restoration framework. All variants share the same
pretrained initialization, restoration generator,
LoRA configuration, training data, degradation
pipeline, reconstruction losses, and common
optimization settings. They differ only in the
additional distribution-level supervision and the
mechanisms used to provide it.
For convergence analysis, we extend training
to examine performance over longer training budgets.

For a fair comparison, we evaluate restoration quality
under approximately matched training budgets,
measured in A100 GPU-hours:
\begin{equation}
\mathrm{GPU\text{-}hours}
=
\frac{N_{\mathrm{step}}t_{\mathrm{step}}N_{\mathrm{GPU}}}{3600}.
\end{equation}

As illustrated in Fig.~\ref{fig:loss_efficiency},
FDIR converges faster, achieves better perceptual
quality, and requires less peak GPU memory than
the VSD- and GAN-based variants.
Table~\ref{tab:matched_budget} further shows that
the full FD loss achieves better perceptual quality
than VSD and GAN under approximately matched
training budgets, outperforming both alternatives
on all evaluated metrics.
These results demonstrate that FDIR provides an
effective and efficient alternative to adversarial
training and score distillation, without requiring
additional trainable discriminators or online
score estimators.

\subsection{Ablation Study}

\subsubsection{Effect of Distribution-Level Supervision.}

We investigate the effectiveness of distribution-level
supervision by comparing the full FD loss in FDIR with
conventional reconstruction and feature-space MSE
objectives. Specifically, we consider three training
objectives: MSE and LPIPS, MSE and LPIPS augmented with
feature-space MSE, and MSE and LPIPS augmented with
the full FD loss. Feature-space MSE directly minimizes
the discrepancy between the SR and paired HQ features
extracted by frozen representation encoders, whereas
the full FD loss performs distribution matching in the
same feature spaces. The Feature MSE variant serves as
a controlled baseline to examine whether the benefits
of distribution-level supervision arise merely from
the pretrained representation spaces rather than the
distribution-matching objective itself. Both variants
employ identical encoders and feature levels to ensure
a fair comparison.

\begin{figure}[t]
    \centering
    \includegraphics[width=0.8\linewidth]{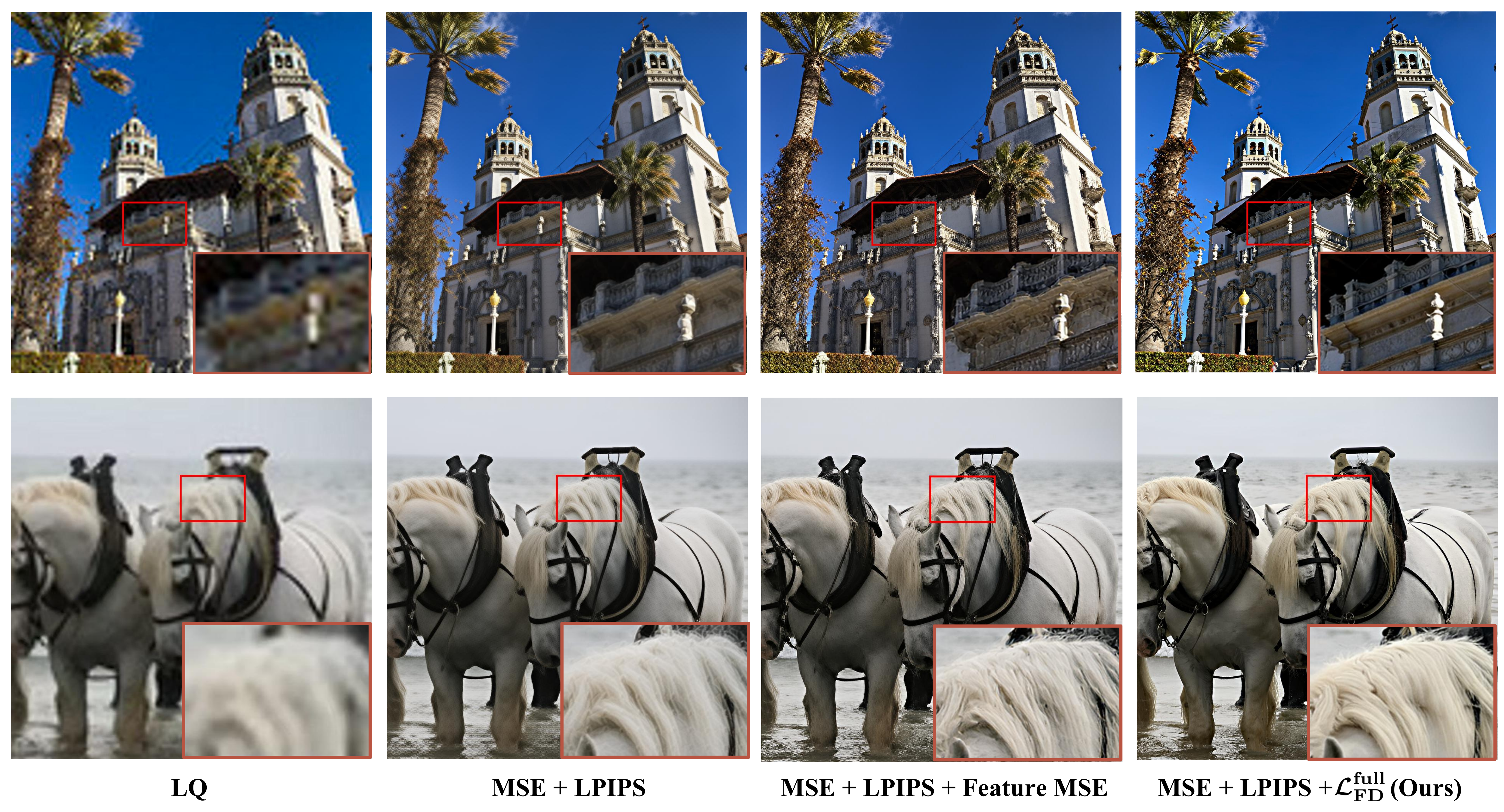}
    \caption{
    Qualitative comparison of reconstruction supervision,
    feature-space MSE, and the full FD loss.
    }
    \label{fig:distribution_supervision}
\end{figure}

As shown in Table~\ref{tab:distribution_supervision},
introducing feature-space MSE improves the quantitative
results over the MSE and LPIPS baseline. However, these
improvements do not necessarily translate into
satisfactory visual quality, as illustrated in
Fig.~\ref{fig:distribution_supervision}. While the
baseline produces over-smoothed textures, feature-space
MSE recovers additional details but still exhibits
unnatural and incoherent textures. In contrast, models
trained with the full FD loss produce more realistic
and coherent details while achieving the best
performance across all reported perceptual metrics.
Since both variants employ the same pretrained
representations, the improvements achieved by the full
FD loss cannot be attributed solely to the
representation spaces themselves. These results
demonstrate the importance of explicitly matching
feature distributions, rather than directly minimizing
feature discrepancies, for high-quality generative
restoration.

\subsubsection{Effect of PatchFD.}

We first evaluate the effectiveness of PatchFD.
As shown in Table~\ref{tab:fd_ablation}, introducing
PatchFD consistently improves restoration performance,
both with and without AdaFD, demonstrating its
complementary benefits to image-level FD matching.

\begin{figure}[t]
    \centering
    \includegraphics[width=0.8\linewidth]{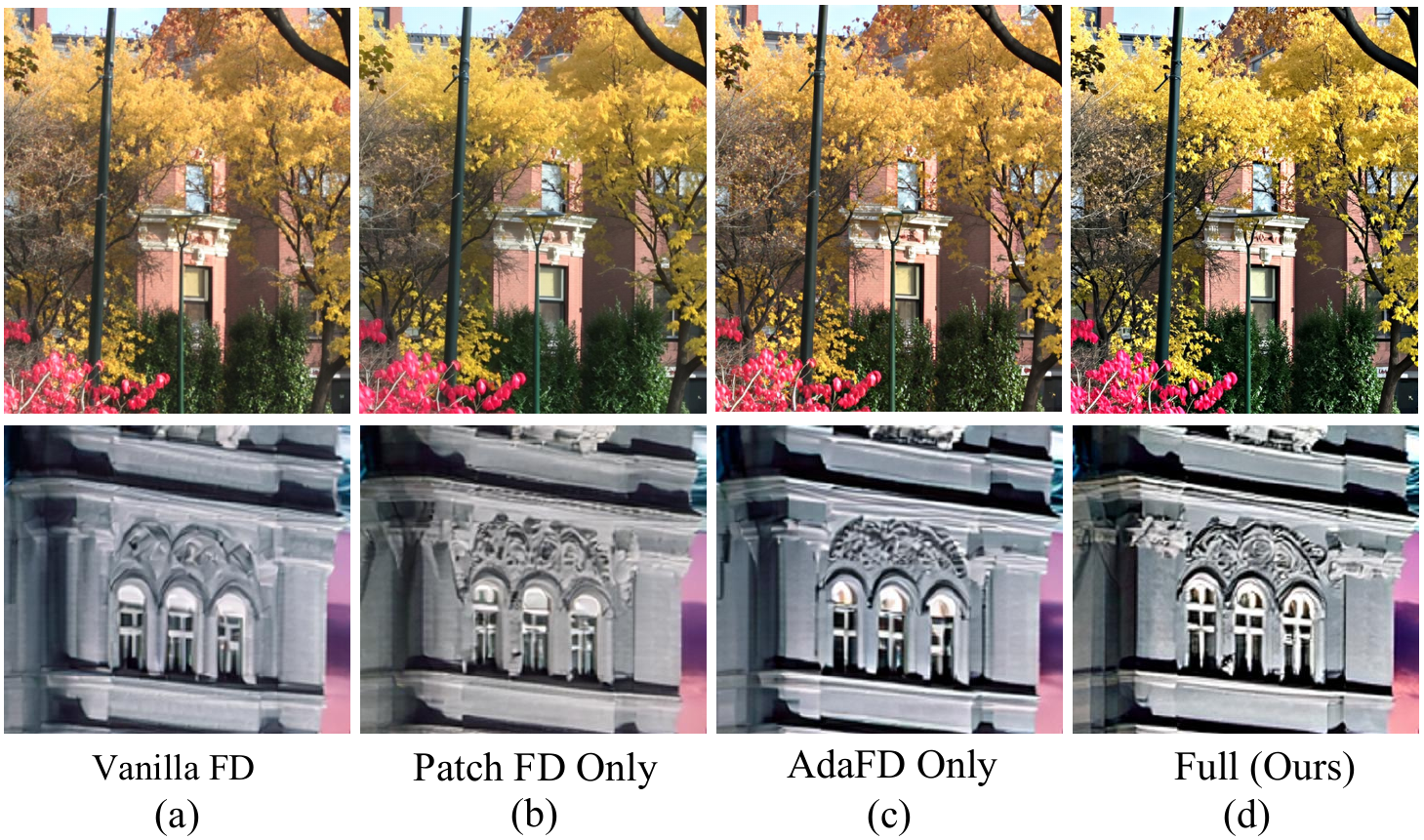}
    \caption{Qualitative ablation of PatchFD and AdaFD.}
    \label{fig:ablation}
\end{figure}

The qualitative comparisons in Fig.~\ref{fig:ablation}
further demonstrate the benefits of patch-level
matching. Compared with image-level FD alone,
incorporating PatchFD recovers clearer foliage
boundaries and more distinct architectural patterns.
Similar improvements are observed when PatchFD is
combined with AdaFD, indicating that patch-level
matching provides complementary supervision for
recovering local textures and structural details.

\begin{table}[t]
\centering
\caption{
Quantitative comparison of different training objectives
on RealPhoto60. Feature MSE and the full FD loss employ
the same frozen representation encoders and feature levels.
}
\label{tab:distribution_supervision}
\resizebox{\linewidth}{!}{
\begin{tabular}{lcccccc}
\toprule
Training Objective
& MUSIQ$\uparrow$
& NIQE$\downarrow$
& CLIP-IQA$\uparrow$
& DeQA$\uparrow$
& Q-Insight$\uparrow$
& RALI$\uparrow$ \\
\midrule
MSE + LPIPS
& 68.669
& 5.107
& 0.677
& 3.947
& 4.051
& 3.935 \\

MSE + LPIPS + Feature MSE
& 71.732
& 4.046
& 0.763
& 4.271
& 4.059
& 3.996 \\

MSE + LPIPS + $\mathcal{L}_{\mathrm{FD}}^{\mathrm{full}}$ (Ours)
& \textbf{74.832}
& \textbf{2.820}
& \textbf{0.788}
& \textbf{4.371}
& \textbf{4.195}
& \textbf{4.165} \\
\bottomrule
\end{tabular}
}
\end{table}

\begin{table}[t]
    \centering
    \caption{
Ablation study of PatchFD and AdaFD on RealPhoto60.
All variants retain image-level FD matching.
The best results are highlighted in bold.
}
    \label{tab:fd_ablation}

    \footnotesize
    \setlength{\tabcolsep}{4pt}
    \renewcommand{\arraystretch}{1.15}

    \resizebox{\linewidth}{!}{
    \begin{tabular}{@{}cccccccc@{}}
        \toprule
        PatchFD
        & AdaFD
        & MUSIQ $\uparrow$
        & NIQE $\downarrow$
        & CLIP-IQA $\uparrow$
        & DeQA $\uparrow$
        & Q-Insight $\uparrow$
        & RALI $\uparrow$ \\
        \midrule

        w/o & w/o
        & 69.077 & 3.743 & 0.680
        & 3.933 & 4.016 & 3.970 \\

        w/ & w/o
        & 72.672 & 3.171 & 0.720
        & 4.291 & 4.158 & 4.098 \\

        w/o & w/
        & 72.941 & 2.946 & 0.738
        & 4.293 & 4.162 & 4.115 \\

        w/ & w/
        & \textbf{74.832}
        & \textbf{2.820}
        & \textbf{0.788}
        & \textbf{4.371}
        & \textbf{4.195}
        & \textbf{4.165} \\

        \bottomrule
    \end{tabular}
    }
\end{table}

\subsubsection{Effect of AdaFD}


We further investigate the effectiveness of AdaFD.
As shown in Table~\ref{tab:fd_ablation}, introducing
AdaFD consistently improves restoration performance,
both with and without PatchFD. The qualitative
comparisons in Fig.~\ref{fig:ablation} also show
that AdaFD produces cleaner structures, more natural
local contrast, and better-preserved restoration
details. These results demonstrate that
instance-adaptive statistics provide additional
benefits beyond conventional Fr\'echet matching
by incorporating input-specific SR--HQ correspondence.

To further examine the role of instance-specific
SR--HQ correspondence, we introduce a
Shuffled-Pair AdaFD variant that randomly permutes
the current HQ targets within each mini-batch
while retaining the historical statistics.
This operation preserves the batch-level HQ
statistics but disrupts the original SR--HQ pairing,
allowing us to assess the importance of
correct instance correspondence.

\begin{figure}[!b]
    \centering
    \includegraphics[
        width=0.8\textwidth
    ]{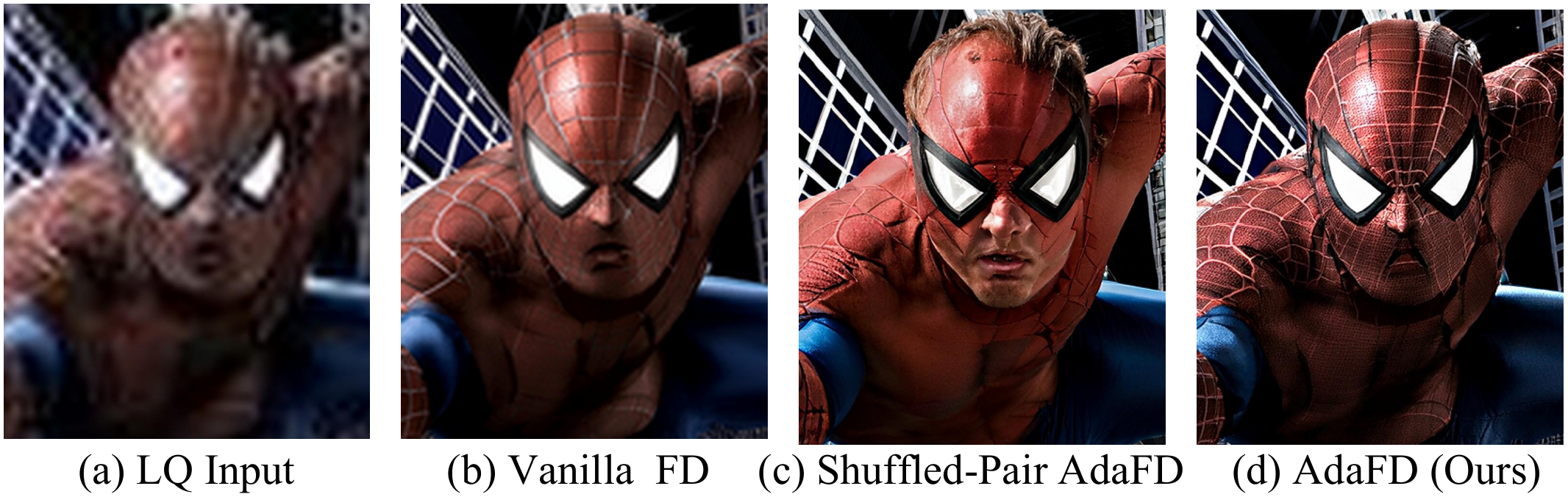}
    \caption{
         Qualitative ablation of AdaFD. Compared with AdaFD, Shuffled-Pair AdaFD exhibits visible deviations in subject appearance and local
    structures, highlighting the importance of preserving the
    original SR--HQ pairing.
    }
    \label{fig:current_pair_ablation}
\end{figure}

As illustrated in
Fig.~\ref{fig:current_pair_ablation},
vanilla FD produces less faithful structures and
textures, while Shuffled-Pair AdaFD exhibits
noticeable deviations in subject appearance
and local structures despite preserving the
batch-level HQ statistics. In contrast, AdaFD
better preserves the original content and
produces more coherent restoration results.
These observations demonstrate the importance
of correct SR--HQ correspondence in
instance-adaptive distribution matching.

\subsection{Further Analysis}

\raggedbottom

\begin{figure}[!t]
    \centering
    \includegraphics[
        width=0.8\linewidth,
        height=0.63\textheight,
        keepaspectratio
    ]{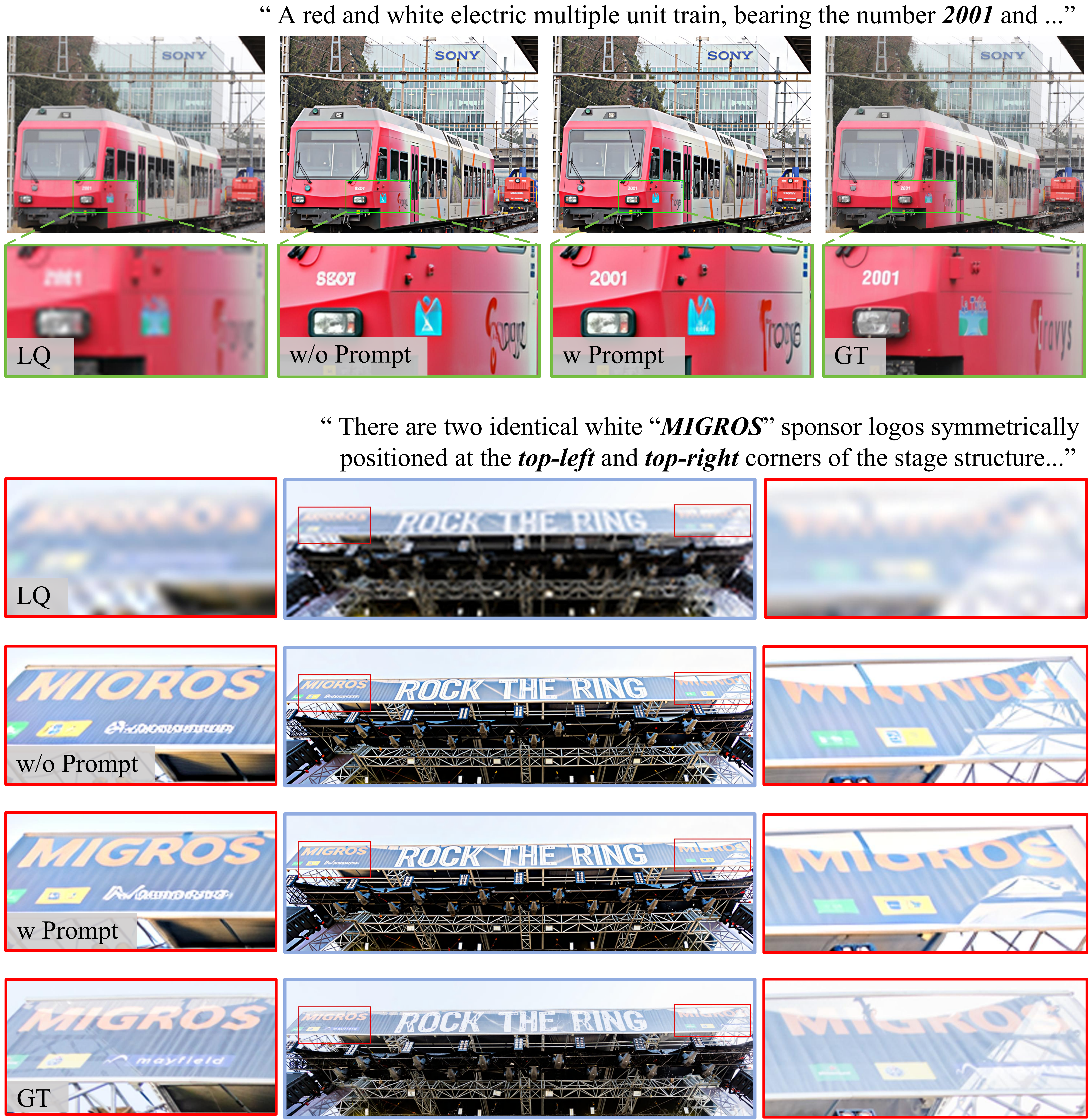}
    \caption{
        Illustration of text-guided correction. Without an explicit prompt,
        textual content in degraded images may be misinterpreted or
        incorrectly restored. Providing a clear semantic prompt helps FDIR
        recover the intended text more faithfully.
    }
    \label{fig:prompt_control}
\end{figure}

\paragraph{Prompt-Guided Restoration}
As introduced in Sec.~\ref{sec:restoration_generator},
FDIR retains the textual conditioning capability of
the pretrained generative backbone.
As shown in Fig.~\ref{fig:prompt_control},
severely degraded textual content may be incorrectly
reconstructed under default restoration settings.
By providing explicit textual prompts, FDIR can
recover the intended content more faithfully,
such as train numbers and signage, while preserving
the surrounding structures.
These results demonstrate that textual guidance
can help resolve semantic ambiguities that are
difficult to infer from degraded visual inputs alone.

\begin{figure}[!t]
    \centering
    \includegraphics[
        width=0.8\linewidth,
        height=0.30\textheight,
        keepaspectratio
    ]{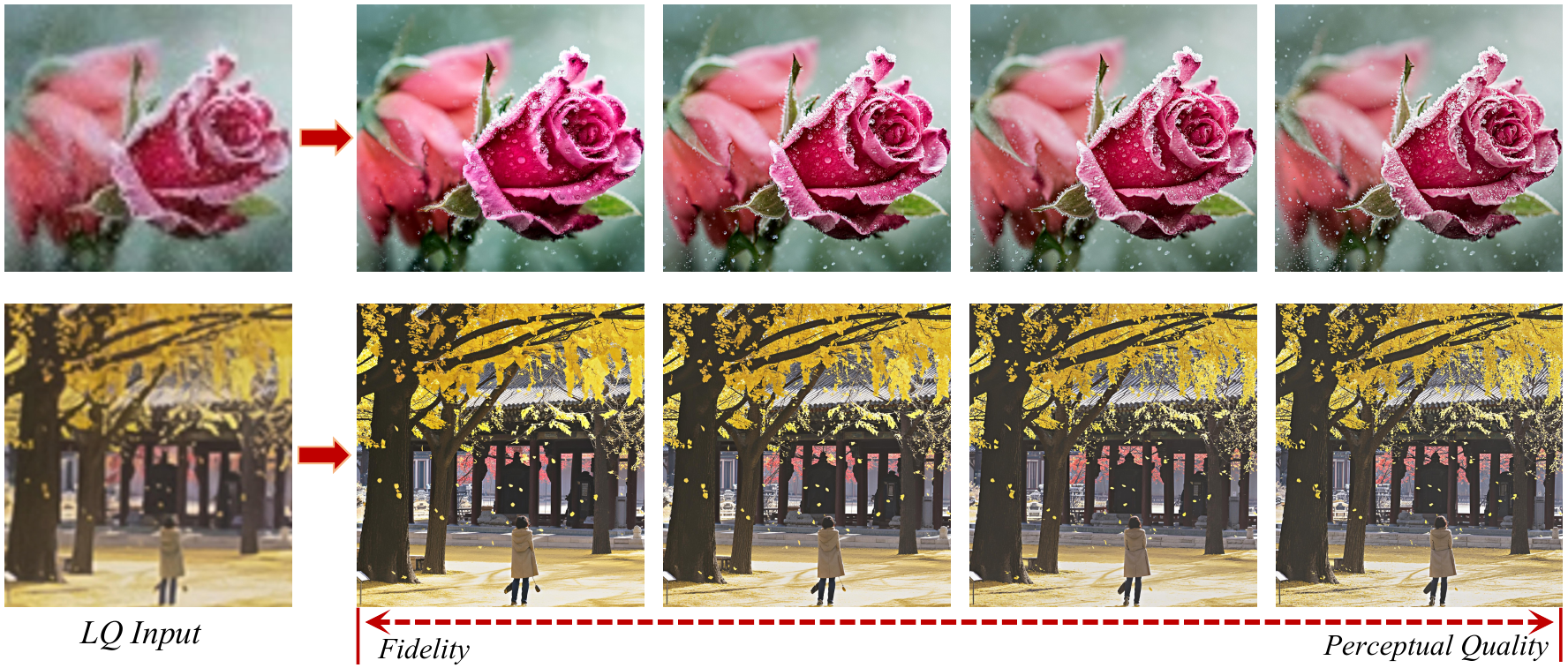}
    \caption{
        Illustration of controlling the fidelity--perceptual quality
        trade-off through noise strength. A higher noise strength suppresses
        compression artifacts and enables richer texture recovery for
        severely degraded inputs.
    }
    \label{fig:fidelity_tradeoff}
\end{figure}

\paragraph{Fidelity--Perception Trade-off}
We further investigate the effect of latent noise
strength on restoration results.
As illustrated in Fig.~\ref{fig:fidelity_tradeoff},
smaller noise strengths encourage greater fidelity
to the LQ input but may retain compression artifacts
and degraded textures. Increasing the noise strength
allows greater generative flexibility, enabling
more effective artifact removal and texture refinement.
However, excessive noise may also introduce deviations
from the original content.
These observations demonstrate that FDIR provides
a flexible trade-off between input fidelity and
perceptual quality without additional training.

\begin{figure}[!t]
    \centering
    \includegraphics[
        width=0.8\textwidth,
        height=1\textheight,
        keepaspectratio
    ]{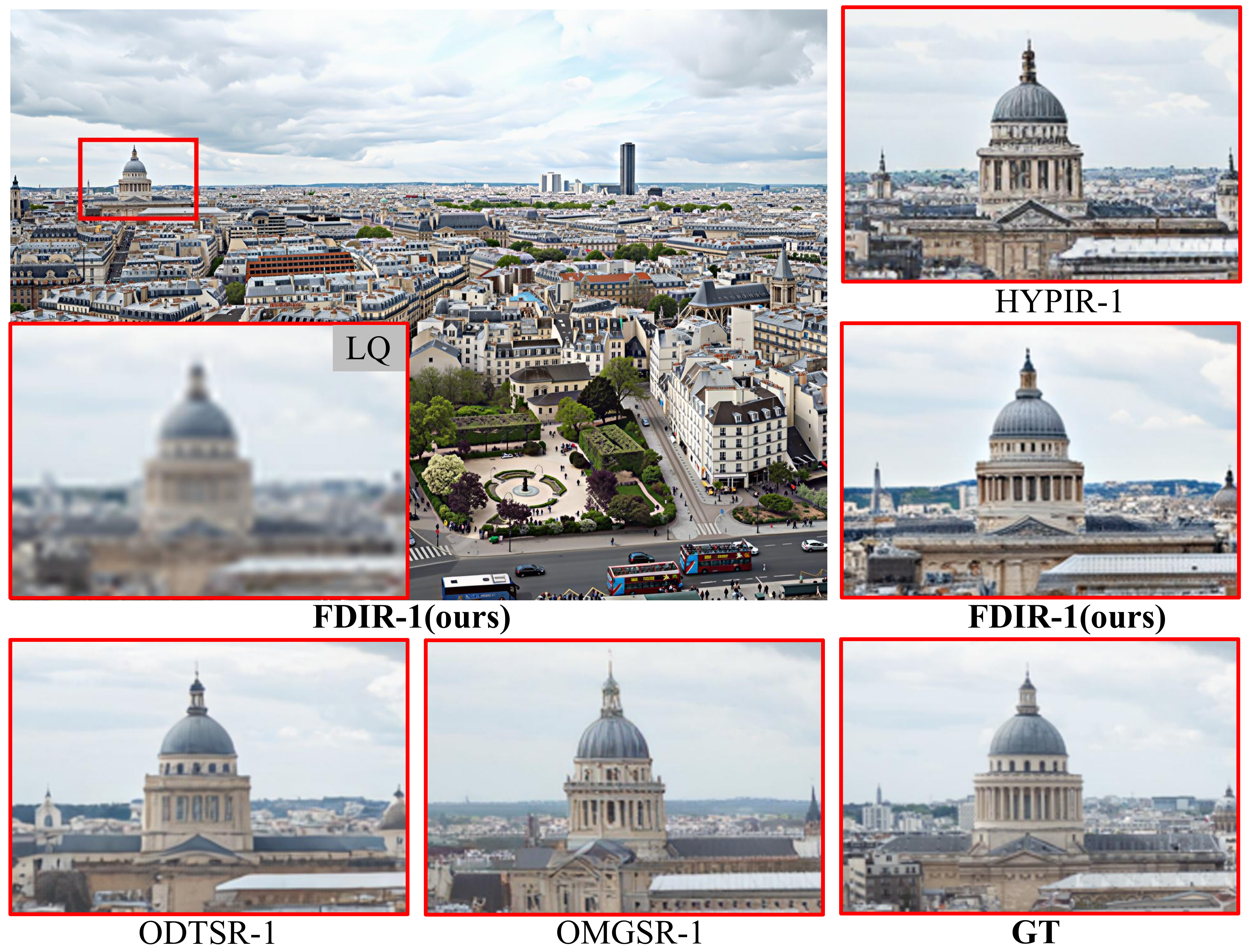}
    \caption{
        Qualitative comparisons with existing
    restoration methods on Test2K.
    }
    \label{fig:test2k}
\end{figure}

\paragraph{High-Resolution Restoration}
We further present qualitative comparisons on
Test2K~\cite{Classsr} to demonstrate FDIR's
restoration capability at higher resolutions.
As shown in Fig.~\ref{fig:test2k}, FDIR produces
visually appealing restoration results at 2K resolution,
preserving coherent structures and fine details.
These results suggest that FDIR can also be applied
to higher-resolution image restoration.

\section{Conclusion and Limitations}

We presented FDIR, a regression-based supervised
framework for single-pass generative image restoration.
By incorporating Fr\'echet-based distribution matching
into paired training, FDIR combines image-level and
patch-level supervision with instance-adaptive statistics.
Extensive experiments on real-world and synthetic
benchmarks demonstrate its strong perceptual restoration
quality and favorable training efficiency.

Despite its effectiveness, FDIR remains limited by the statistical and representational capacity of its supervision. Although PatchFD and AdaFD provide local and pair-specific supervision, Fréchet matching still relies on first- and second-order feature statistics and does not explicitly enforce pixel-wise correspondence. Moreover, frozen encoders may not fully capture all restoration-relevant perceptual attributes.

Nevertheless, we view this work as an initial exploration
of regression-based generative restoration. Rather than
providing a definitive solution to generative restoration
training, it demonstrates the potential of directly
computable distribution-level supervision as a simpler
training paradigm. We hope this work provides a promising
direction for further research into simpler and more
effective generative restoration methods.

\bibliographystyle{plainnat}
\footnotesize
\bibliography{main}

\clearpage
\normalsize
\appendix
\raggedbottom
\numberwithin{equation}{section}
\numberwithin{figure}{section}
\numberwithin{table}{section}
\section{Practical Implementation}
\label{app:implementation}

\subsection{Restoration Generator}
\label{sec:restoration_generator}

We adopt FLUX.2-klein-base-4B as the pretrained backbone for
single-step image restoration. To adapt its generative prior to
restoration, we retain its textual conditioning capability and introduce
controllable noise augmentation in the LQ latent space, as illustrated
in Fig.~\ref{fig:restoration_generator}.

\paragraph{Textual Prompts}
The pretrained FLUX.2 model inherently supports textual conditioning,
which provides semantic guidance for restoring ambiguous or missing
image details. Following the captioning strategy of prior
work~\cite{SUPIR,HYPIR}, we employ LLaVA~\cite{LLAVA} to generate textual
descriptions for the training images. At inference time, LLaVA
automatically generates prompts from LQ inputs, enabling restoration
without manual intervention. Users can also provide customized prompts
to guide the reconstruction of ambiguous content.

\paragraph{Latent Noise Augmentation}
To balance restoration fidelity and generative flexibility, we
introduce Gaussian noise into the encoded LQ latent. The noise strength
controls the extent to which the restoration process relies on the
observed input. A smaller noise strength preserves more input
information, whereas a larger strength allows greater generative
flexibility, which can be beneficial for severely degraded regions but
may also introduce deviations from the input. During training, we
randomly sample the noise strength from $[0.2,0.8]$. At inference time,
it can be adjusted to control the fidelity--perception trade-off.

\paragraph{Overall Restoration Process}
Given an LQ image $x_{\mathrm{LQ}}$ and a textual prompt $y$, the
restoration process is formulated as
\begin{align}
z_{\mathrm{LQ}} &= \mathcal{E}(x_{\mathrm{LQ}}),
& c &= \mathcal{T}(y), \\
z'_{\mathrm{LQ}} &= z_{\mathrm{LQ}}+\tau\epsilon,
& v &= G(z'_{\mathrm{LQ}},t^*,c), \\
z_{\mathrm{SR}} &= z'_{\mathrm{LQ}}-\Delta t v,
& x_{\mathrm{SR}} &= \mathcal{D}(z_{\mathrm{SR}}).
\end{align}
Here, $\mathcal{E}$ and $\mathcal{D}$ denote the frozen VAE encoder and
decoder, respectively, and $\mathcal{T}$ denotes the frozen text
encoder. The Gaussian noise $\epsilon\sim\mathcal{N}(0,I)$ is scaled by
$\tau$. The FLUX.2 backbone $G$ takes the fixed timestep $t^*$ and
textual condition $c$ to predict the velocity $v$, which is used to
obtain the restored latent through a single update.

\subsection{Representation Encoders}

We employ SigLIP SO400M/16, MAE ViT-L/16, and MANIQA. SigLIP and
MAE operate at input resolutions of $512\times512$ and $224\times224$,
respectively. For MANIQA, we extract intermediate features after the
second Swin Transformer stage and before the quality regression heads.
For each encoder, we extract both image-level and patch-level features
for FD supervision. The selection of these three complementary
representation spaces is analyzed below.
\label{app:encoder_selection}

Since the FD loss performs distribution matching in frozen feature
spaces, its effectiveness depends on the representations used for
supervision. Different pretrained encoders emphasize different visual
attributes, and supervision from a single representation space may not
adequately constrain all aspects of image restoration. Beyond commonly
used semantic and reconstruction-based encoders, we further consider
image quality assessment models, whose intermediate features are
sensitive to perceptual quality and may provide complementary
supervision.

\begin{center}
\centering
\includegraphics[width=0.78\linewidth]{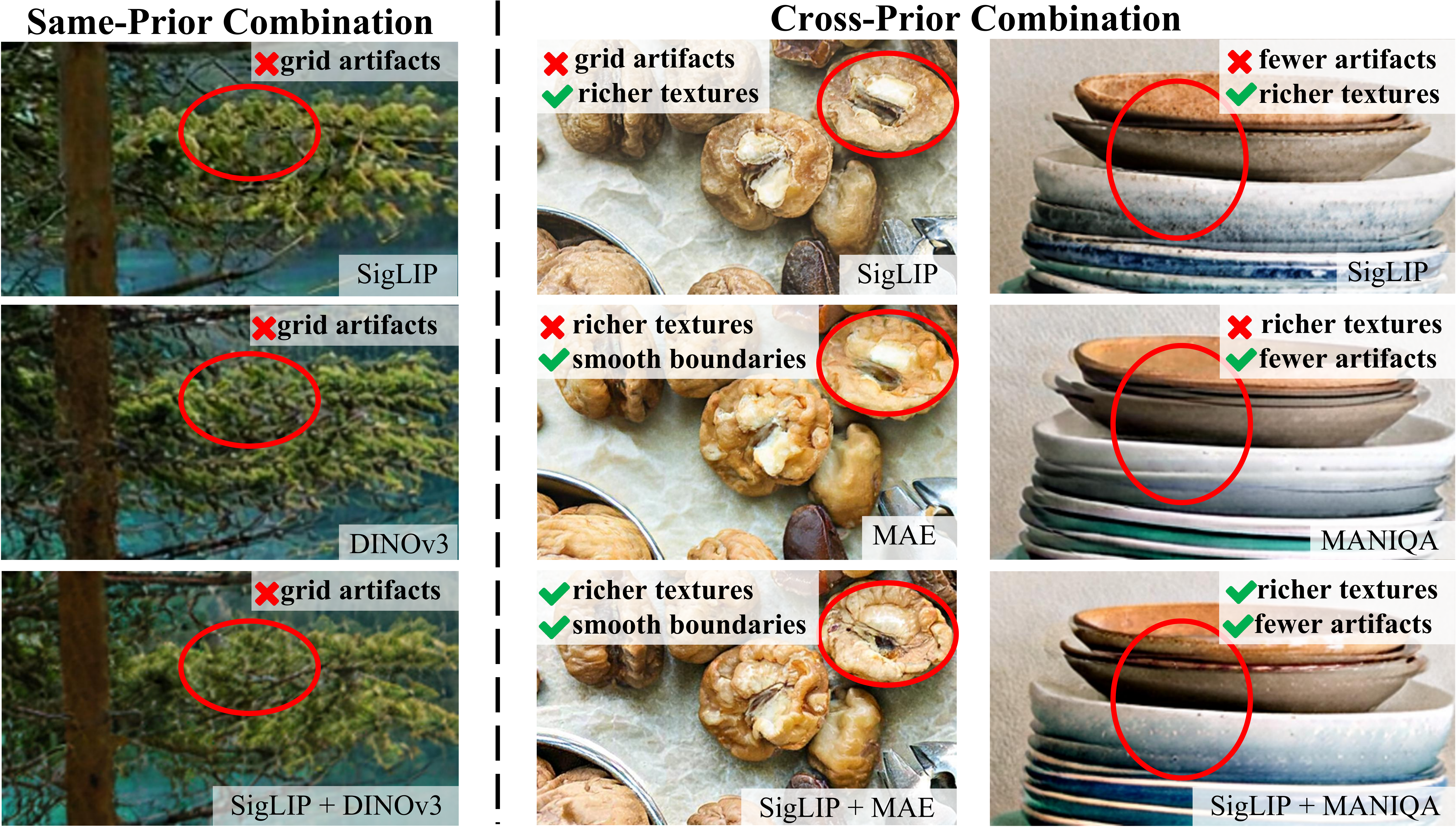}
\captionof{figure}{Qualitative comparison of individual representation encoders
and their combinations, illustrating the complementary effects of
different representation spaces.}
\label{fig:compare01}
\end{center}

\begin{center}
\captionof{table}{Quantitative comparison of representation encoders and their
combinations on RealLQ250. Bold indicates the best results within each
category; $\dagger$ marks the encoders used in the final configuration.
C-IQA denotes CLIP-IQA, Q-Ins. denotes Q-Insight, and MAN. denotes MANIQA.}
\label{tab:representation_selection}
\centering
\footnotesize
\setlength{\tabcolsep}{1.2pt}
\renewcommand{\arraystretch}{1.08}
\begin{tabular*}{\linewidth}{@{\extracolsep{\fill}}llccccc@{}}
\toprule
\textbf{Prior Type} & \textbf{Encoder} & \textbf{NIQE}$\downarrow$
& \textbf{C-IQA}$\uparrow$ & \textbf{DeQA}$\uparrow$
& \textbf{Q-Ins.}$\uparrow$ & \textbf{RALI}$\uparrow$ \\
\midrule
\multicolumn{7}{c}{\textbf{Individual Representation Encoders}} \\
\midrule
\multirow{4}{*}{\textit{Semantic}}
& CLIP & 4.054 & 0.690 & 4.025 & 3.983 & 3.953 \\
& DINOv2 & 3.696 & 0.679 & 4.032 & 3.985 & 3.915 \\
& DINOv3 & 3.755 & 0.665 & 4.047 & 3.997 & 3.904 \\
& SigLIP$^{\dagger}$ & \textbf{3.471} & \textbf{0.712}
& \textbf{4.053} & \textbf{4.007} & \textbf{3.958} \\
\midrule
\textit{Reconstruction} & MAE$^{\dagger}$
& 3.698 & 0.701 & 4.146 & 4.035 & 4.055 \\
\midrule
\multirow{2}{*}{\textit{Quality-aware}}
& MUSIQ & \textbf{3.536} & 0.697 & 4.031 & 4.018 & 3.997 \\
& MAN.$^{\dagger}$ & 3.648 & \textbf{0.710} & \textbf{4.082}
& \textbf{4.046} & \textbf{4.018} \\
\midrule
\multicolumn{7}{c}{\textbf{Representation Combinations}} \\
\midrule
\textit{Same-prior} & SigLIP+DINOv3
& 3.556 & 0.673 & 3.974 & 3.971 & 3.899 \\
\midrule
\multirow{3}{*}{\textit{Cross-prior}}
& SigLIP+MAE & 3.277 & 0.737 & 4.130 & 4.065 & 4.024 \\
& SigLIP+MAN. & 3.120 & 0.731 & 4.121 & 4.053 & 4.034 \\
& \textbf{SigLIP+MAE+MAN.} & \textbf{3.105} & \textbf{0.756}
& \textbf{4.169} & \textbf{4.087} & \textbf{4.046} \\
\bottomrule
\end{tabular*}
\end{center}

As illustrated in Fig.~\ref{fig:compare01}, different representation
encoders exhibit distinct restoration characteristics. SigLIP and
DINOv3 recover rich textures but retain grid-like artifacts in foliage.
In contrast, MAE produces smoother boundaries at the expense of some
fine-grained details, while MANIQA reduces visible artifacts but may
weaken textures. These observations suggest that a single
representation space may leave certain restoration discrepancies
insufficiently constrained.

Combining these encoders further highlights their complementary
effects. SigLIP+DINOv3 does not noticeably alleviate the grid-like
artifacts, whereas SigLIP+MAE improves structural continuity while
retaining rich textures. Similarly, SigLIP+MANIQA reduces artifacts
while preserving texture details. These observations suggest that
combining complementary representation spaces is more beneficial than
combining similar encoders.

The quantitative results in Table~\ref{tab:representation_selection}
further support these observations. Compared with individual encoders
and the same-prior combination, cross-prior combinations generally
provide a more balanced performance across different perceptual
metrics. We therefore adopt
$\Phi=\{\mathrm{SigLIP},\mathrm{MAE},\mathrm{MANIQA}\}$ as the final
set of frozen representation encoders, covering semantic,
reconstruction-oriented, and quality-aware representations,
respectively.

\FloatBarrier

\subsection{Training Configuration}

We train FDIR for 5K iterations, optimizing the LoRA parameters of the
FLUX.2 backbone using Adam with a learning rate of $1\times10^{-5}$ and
a global batch size of 64. The LoRA rank is set to 64. We use a fixed
restoration timestep $t^*=500$ and sample the noise augmentation
strength $\tau\sim\mathcal{U}(0.2,0.8)$ during training. At inference
time, the noise strength is fixed at $\tau=0.2$ for all quantitative
evaluations.

The loss weights are set to $\lambda_{\mathrm{MSE}}=1$,
$\lambda_{\mathrm{LPIPS}}=1$, and $\lambda_{\mathrm{FD}}=500$. For the
instance-adaptive Fr\'echet statistics, the historical update
coefficient $\beta$ and the instance interpolation coefficient $\alpha$
are both set to $0.999$. Training is performed on $8\times$ NVIDIA A100
GPUs and takes approximately 8 hours.

\subsection{Training Data and Evaluation Protocol}

The overall training data includes 200K high-quality images with text
descriptions. Following the captioning protocol of SUPIR~\cite{SUPIR},
we employ LLaVA~\cite{LLAVA} to generate textual descriptions for the
training images, which serve as text conditions during training. Each
HQ image is randomly cropped to $1024\times1024$ and resized to
$512\times512$. The corresponding LQ image is synthesized using the
two-stage degradation pipeline of Real-ESRGAN~\cite{RealESRGAN}.

For synthetic evaluation, we use DIV2K-Val~\cite{Div2k}, where LQ
inputs are generated through $\times4$ downsampling followed by the
Real-ESRGAN degradation pipeline. For real-world evaluation, we employ
RealPhoto60~\cite{SUPIR} and RealLQ250~\cite{RealLQ250}, which contain
real-world LQ images without paired HQ references. For DIV2K-Val, we
report PSNR, SSIM, LPIPS, and DISTS. For the unpaired real-world
benchmarks, we report MUSIQ, NIQE, CLIP-IQA, DeQA, Q-Insight, and RALI.

\FloatBarrier

\section{Hyperparameter Analysis}
\label{app:hyperparameters}

\paragraph{Effect of $\lambda_{\mathrm{FD}}$}
We study the sensitivity to the full FD loss weight by varying
$\lambda_{\mathrm{FD}}\in\{100,300,500,700\}$ while fixing
$\alpha=0.999$. As shown in Table~\ref{tab:lambda_sensitivity}, the
overall performance remains relatively stable across a wide range of
loss weights.

\begin{center}
\centering
\captionof{table}{Sensitivity to the full FD loss weight on RealPhoto60 with
$\alpha=0.999$.}
\label{tab:lambda_sensitivity}
\resizebox{0.8\linewidth}{!}{%
\begin{tabular}{ccccccc}
\toprule
$\lambda_{\mathrm{FD}}$ & MUSIQ$\uparrow$ & NIQE$\downarrow$
& CLIP-IQA$\uparrow$ & DeQA$\uparrow$ & Q-Insight$\uparrow$
& RALI$\uparrow$ \\
\midrule
100 & 73.441 & 3.196 & 0.781 & 4.299 & 4.151 & 4.066 \\
300 & 73.882 & 2.937 & 0.777 & 4.301 & 4.152 & 4.114 \\
500 & \textbf{74.832} & 2.820 & \textbf{0.788} & \textbf{4.371}
& \textbf{4.195} & \textbf{4.165} \\
700 & 74.216 & \textbf{2.726} & 0.772 & 4.307 & 4.160 & 4.127 \\
\bottomrule
\end{tabular}}
\end{center}

\paragraph{Effect of $\alpha$}
Since the historical statistics are treated as constants during
back-propagation, gradients from the full FD loss flow through only the
current SR statistics. By the chain rule, the explicit gradient-scale
factor introduced by the interpolation is
$\lambda_{\mathrm{FD}}(1-\alpha)$. Directly varying $\alpha$ under a
fixed loss weight would therefore change both the composition of the
temporary statistics and the effective FD gradient magnitude.

To isolate the effect of the history--current composition, we jointly
adjust $\alpha$ and $\lambda_{\mathrm{FD}}$ such that
$\lambda_{\mathrm{FD}}(1-\alpha)=0.5$, matching the default setting.
This approximately controls the explicit gradient rescaling while
varying the contribution of the current observation to the temporary
statistics.

\begin{center}
\centering
\captionof{table}{Sensitivity to $\alpha$ under matched explicit gradient
scaling on RealPhoto60.}
\label{tab:alpha_sensitivity}
\resizebox{0.8\linewidth}{!}{%
\begin{tabular}{ccccccc}
\toprule
$\alpha$ & MUSIQ$\uparrow$ & NIQE$\downarrow$
& CLIP-IQA$\uparrow$ & DeQA$\uparrow$ & Q-Insight$\uparrow$
& RALI$\uparrow$ \\
\midrule
0.9 & 72.679 & 3.251 & \underline{0.772} & 4.300 & 4.130 & 4.084 \\
0.99 & 73.369 & 3.103 & 0.756 & \underline{4.356} & 4.154 & 4.093 \\
0.999 & \textbf{74.832} & \textbf{2.820} & \textbf{0.788}
& \textbf{4.371} & \textbf{4.195} & \textbf{4.165} \\
0.9999 & \underline{73.538} & \underline{2.987} & 0.757 & 4.336
& \underline{4.178} & \underline{4.143} \\
\bottomrule
\end{tabular}}
\end{center}

\FloatBarrier






\end{document}